\documentclass[10pt,twocolumn,letterpaper]{article}

\usepackage{cvpr}              

\usepackage[dvipsnames]{xcolor}

\definecolor{cvprblue}{rgb}{0.21,0.49,0.74}
\usepackage[pagebackref,breaklinks,colorlinks,citecolor=cvprblue]{hyperref}
\usepackage{wrapfig}
\usepackage{multirow}
\usepackage{algorithm}
\usepackage{algpseudocode}
\usepackage{stfloats} 
\usepackage{amsmath}
\usepackage{amssymb}
\usepackage{enumitem}
\usepackage{pifont}
\def\paperID{275} 
\def\confName{3DV\xspace}
\def\confYear{2026\xspace}

\title{Hierarchical Space Partition for Surface Reconstruction}

\author{Minjie Tang\\
Independent Researcher\\
\texttt{tangminjie.ai@gmail.com}
\and
Xiangfei Li*\\
Huazhong University of Science and Technology\\
\texttt{lixiangfei@hust.edu.cn}
\thanks{Corresponding author.}}

\makeatletter
\apptocmd{\@maketitle}{%
  \vspace*{-12pt}%
  \centerline{\smash{\normalsize\href{https://hsr-3dv.github.io/}{\texttt{https://hsr-3dv.github.io/}}}}%
  \vspace*{12pt}%
}{}{}
\makeatother

\begin{document}
\maketitle
\begin{abstract}
Generating compact polygonal models from point clouds is a key problem in 3D vision and computer graphics. However, due to inherent limitations of LiDAR scanning (e.g. range constraints and occlusions), critical scene information is often missing, leading to degraded reconstruction accuracy. To address this, we propose a plane assembling strategy that effectively recovers missing details while maintaining model compactness. We classify all the planes extracted from the scene into three categories: highly visible, barely visible, and invisible. The invisible planes, which are recovered by scene structure analysis, indicate the missing details. The three types of planes correspond to the three growth priorities. Each plane grows according to the priority level, and the space is partitioned progressively, that is, the hierarchical partition. Subsequently, we generate a watertight polygonal mesh from the partition via a min-cut-based optimization. Finally, comparisons on public datasets show the effectiveness and superiority of our method against mainstream approaches.
\end{abstract}

\section{Introduction}
\label{sec:introduction}

Surface reconstruction is a crucial topic in the fields of 3D vision and computer graphics.
This problem can be formulated as follows: extracting semantic, structural, and topological information from dense, unordered, and redundant point cloud data (typically acquired via LiDAR scanning) or triangular meshes (often generated by Multi-View Stereo, MVS) to produce high-precision, watertight, and lightweight polygonal surface models~\cite{KSR:2020}.
Such models have significant applications in the domains of autonomous driving, robotics, augmented reality/virtual reality, digital twin, and building information modeling (BIM).

Plane assembly is a popular paradigm in contemporary surface reconstruction benchmarks. This approach partitions 3D space using detected planar shapes from input data and subsequently selects specific polygonal facets or polyhedral cells from the partitioned structure to construct simplified mesh models. The method demonstrates multiple advantages: 1) it is broadly applicable due to the prevalence of planar structures in man-made environments; 2) it exhibits some robustness to missing scan data; 3) it provides favorable geometric properties such as surface watertightness, orientability, and convexity of polyhedral elements. However, most existing methods adopt a homogeneous treatment for all detected planes during spatial partitioning, assigning uniform expansion factors~\cite{PolyFit,fang2021structure} or isotropic growth speed~\cite{KSR:2020}. This method inevitably leads to considerable redundancy in spatial partitioning~\cite{sulzer2024concise}. Besides, extending existing planar entities shows limited effectiveness in completing missing regions in scanning scenes~\cite{KSR:2020}.

To address these limitations, we propose a shape-assembling method that yields lighter yet meaningful 3D partitions and can accurately restore the local details of the incomplete scene. Our first contribution is to design a hierarchical space partition strategy based on a kinetic data structure \cite{guibas2018kinetic, KSR:2020}. Unlike the previously described scheme, we divide planes into three different categories based on visibility (highly visible, barely visible, and invisible), corresponding to three levels. Planes at the current level will partition further based on the outcome only after planes at the immediately higher level have completed their spatial partition (i.e. hierarchical partition). At the same time, each plane is given a different growth speed according to its visibility. This hierarchical strategy produces lightweight and semantically meaningful 3D partitions, with substantially reduced algorithmic complexity compared to other methods. Our secondary contribution is to recover missing planes for models. In contrast to the saturated additions of \textit{ghost planes} proposed in \cite{chauve2010robust}, we first use planar boundary segments and planes' intersection lines to detect singular segments, that is segments not contained by any intersection lines. The missing planes are then fitted through these segments. Experiments show that this strategy accurately recovers the missing details of the scanned model without introducing redundant structures. Finally, we evaluate our approach against state-of-the-art methods across datasets varying in complexity, size, and data collection.

\section{Related Work}
\label{sec:prior}

Polygonal mesh reconstruction is a fundamental problem in computer vision and computer graphics. Prior studies can be grouped into three aspects: i) space partition and plane assembly, ii) scene decomposition and topology analysis, and iii) mesh simplification and geometry approximation.

\textbf{Space Partition and Plane assembly.}
Such methods first utilize extracted geometric primitives (e.g. planes) to partition the 3D space. This space is typically the bounding box of the scene data, and it is divided into subregions such as polyhedral cells~\cite{boulch2014piecewise, mura2016piecewise, langlois2019surface} or polygonal facets~\cite{fang2020connect, fang2021structure, PolyFit}. The primary limitation of these partitioning approaches is their $\text{O}(n^3)$ complexity, resulting in high computational/memory costs. The resulting massive polyhedral cells also create a complex optimization space for subsequent surface reconstruction. 
To address this issue, \cite{chauve2010robust} proposed a two-level hierarchical scheme that first partitions space into super-cells and further subdivides only those intersected by planar primitives. This approach also introduced ghost primitives (i.e. vertical and orthogonal planes) to mitigate data incompleteness issues. 
More recently, \cite{KSR:2020} adopted a kinetic data structure, \cite{chen2022} generated a cell complex via adaptive space partitioning, and \cite{sulzer2024concise} introduced a compact plane arrangement to further avoid excessive subdivision. 
Following the space partitioning stage, the surface extraction is achieved either by classifying polyhedral cells as inside or outside the object \cite{verdie2015lod, li2016manhattan, oesau2014indoor}, or by solving a constrained integer programming formulation~\cite{fang2020connect, fang2021structure, PolyFit} to select polygonal facets.

\textbf{Scene decompose and Topology analysis.}
The second category of methods first leverages semantic or structural information to decompose the reconstruction problem into sub-problems focused on reconstructing individual elements. 
These reconstructed elements are then merged using their topological connections to produce the final model. Early approaches \cite{chen2008architectural, schindler2011classification, van2011shape} for planar-faced architectures decomposed scenes into planar regions and reconstructed models by extracting vertices, edges, and facets from the regions' connectivity graph. 
Although the method is efficient, graph inaccuracies often lead to incomplete or erroneous results. Further advancing this approach, methods like those in~\cite{ochmann2016automatic, han2021vectorized, huang2023arrangementnet, xiang2024efficient} decomposed the scene into semantic categories (e.g., ground, ceiling, columns, walls) and employed distinct reconstruction strategies for each category. This semantic decomposition is typically data-driven, potentially leading to semantic inaccuracies when applied to novel scenes. 
To circumvent this issue, \cite{fang2021structure} relied on predefined distance thresholds to classify scene clusters into structural and non-structural objects, which were then reconstructed separately. 
More recently, \cite{pan2025building} introduced a novel inside/outside view analysis to identify 3D structures, then generated models with varying levels of detail. 

\textbf{Mesh simplification and Geometry approximation.}
Instead of directly generating compact polygonal models, extensive work in computer graphics has focused on simplifying dense meshes by reducing their facet counts \cite{li2021feature, chen2023robust} or imposing geometric assumptions for the output surface \cite{gao2022low}. \cite{garland1997surface} proposed iterative vertex-pair contraction using quadric error metrics; \cite{cohen2004variational} performed face clustering guided by geometric proxies; and \cite{salinas2015structure} developed edge-collapse decimation with planar proxy preservation. While these mesh reduction techniques can yield comparably compact models, they often result in polygonal structures with less regularity and structural coherence than direct generation methods.
\begin{figure*}[t]
  \centering
  \includegraphics[width=0.99999\textwidth]{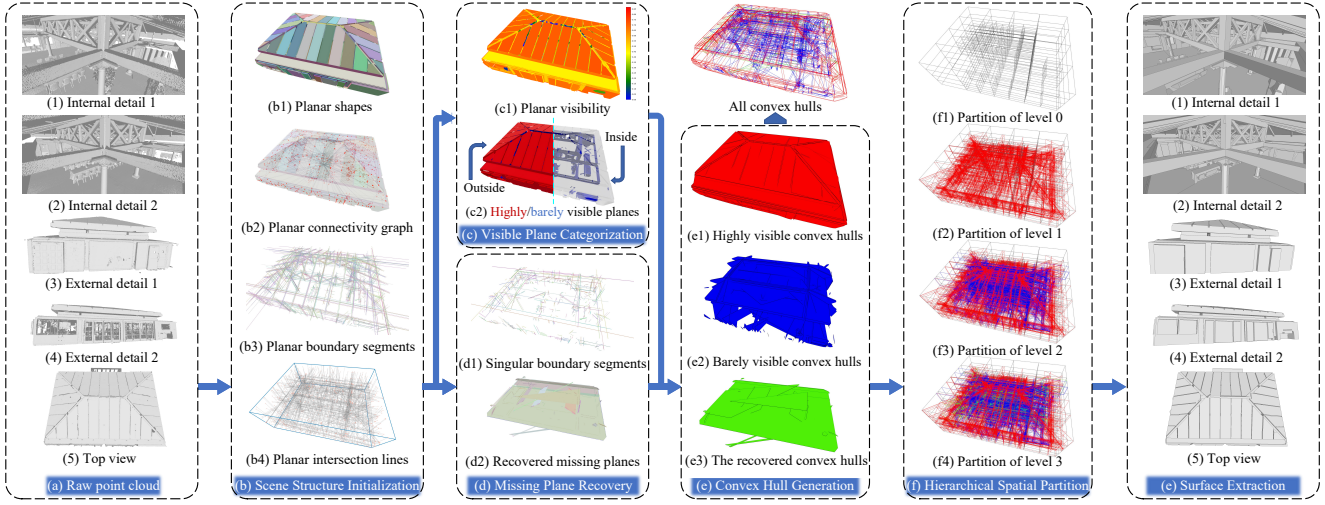}
  \caption{Overview. Our algorithm takes point cloud (a) as the input and first performs the scene structure analysis (b), which mainly includes: planes detection (b1), planar adjacency analysis (b2), boundary segments fitting (b3) and planar intersection lines calculation (b4). Subsequently visible planes categorization (c) is performed: including visible calculation (c1) and highly/barely visible labelling (c2). Meanwhile, we recover missing planes (d) through fitting the singular segments (i.e., the segments that are not contained by any intersection lines (d1)). After that, the 3D bounding box is hierarchically partitioned (f) using the different categories of primitives (i.e., convex hulls of highly visible planes (e1), barely visible planes (e2) and invisible (missing) planes (e3)). Finally, based on the inside/outside labeling of polyhedral cells, a concise, watertight polygonal mesh is generated (f).}
  \label{fig:2}
\end{figure*}

\section{Method}
\label{sec:method}
Taking a point cloud as input, our algorithm outputs a watertight and intersection-free polygonal mesh. Figure \ref{fig:2} illustrates the workflow of the proposed method. We first analyze the scene structure to extract geometric and topological information, which mainly includes planar primitives, intersection lines, the adjacency graph and boundary segments. Then we distinguish the highly/barely visible planes by the visibility. Meanwhile, in order to deal with the possible loss of scanning data, missing planes are recovered by fitting boundary segments of primitives. This type of planes is classified as invisible planes. Next, space partition is accomplished by assigning different levels and expansion speed to planes. Finally, a polygonal surface mesh is reconstructed by classifying polyhedral cells as interior or exterior via min-cut optimization. 

\subsection{Scene Structure Analysis}
Planar primitives are initially extracted from the input point cloud using Random Sample Consensus (RANSAC)-based shape detection methods \cite{rabbani2006segmentation, schnabel2007efficient, cgal:Shape-Detection}. Subsequently, an \( \alpha \)-shape algorithm~\cite{cgal:Alpha-Shapes} is employed to approximate the boundary of each detected primitive, as shown in Figure \ref{fig:2}(b1). Then we project boundary points onto its plane, and calculate the 2D normal vector of each point using principal component analysis. Thereafter, 2D segments are fitted using region growing~\cite{cgal:Shape-Detection}. To mitigate missing data and noise, we adopt the regularization process of~\cite{han2021vectorized} to obtain cleaner and more accurate segments. The 2D segments and the corresponding 2D normal vectors are then back-projected into 3D space to obtain the final set of 3D segments, as shown in Figure \ref{fig:2}(b3).

Using all primitives, an adjacency graph \( \mathcal{G} \) is constructed by identifying the neighboring relationships between them, as illustrated in Figure \ref{fig:2}(b2). Two planes are defined as adjacent if at least two inlier points, one from each primitive, are mutual neighbors in the k-nearest neighbor graph of the input points~\cite{fang2020connect}. Moreover, for any pair of non-parallel primitives \( i \) and \( j \), we compute their intersection line \( L_{ij} \). The set of all such lines is denoted as \( \mathcal{L} \), as in Figure \ref{fig:2}(b4).

\subsection{Visible Plane Categorization}
In this section, we model \( \alpha \)-shapes as uniformly luminous objects to compute individual visibility ratios of primitives. Then, Markov random field (MRF) optimization method is used to allocate highly/barely visible labels for planes.

\textbf{Planar visibility.} 
For each planar primitive \( P_i \), we use 
\setlength{\intextsep}{3pt}
\setlength{\columnsep}{5pt}
\begin{wrapfigure}{r}{0.2\textwidth} 
    \centering
    \includegraphics[width=0.95\linewidth]{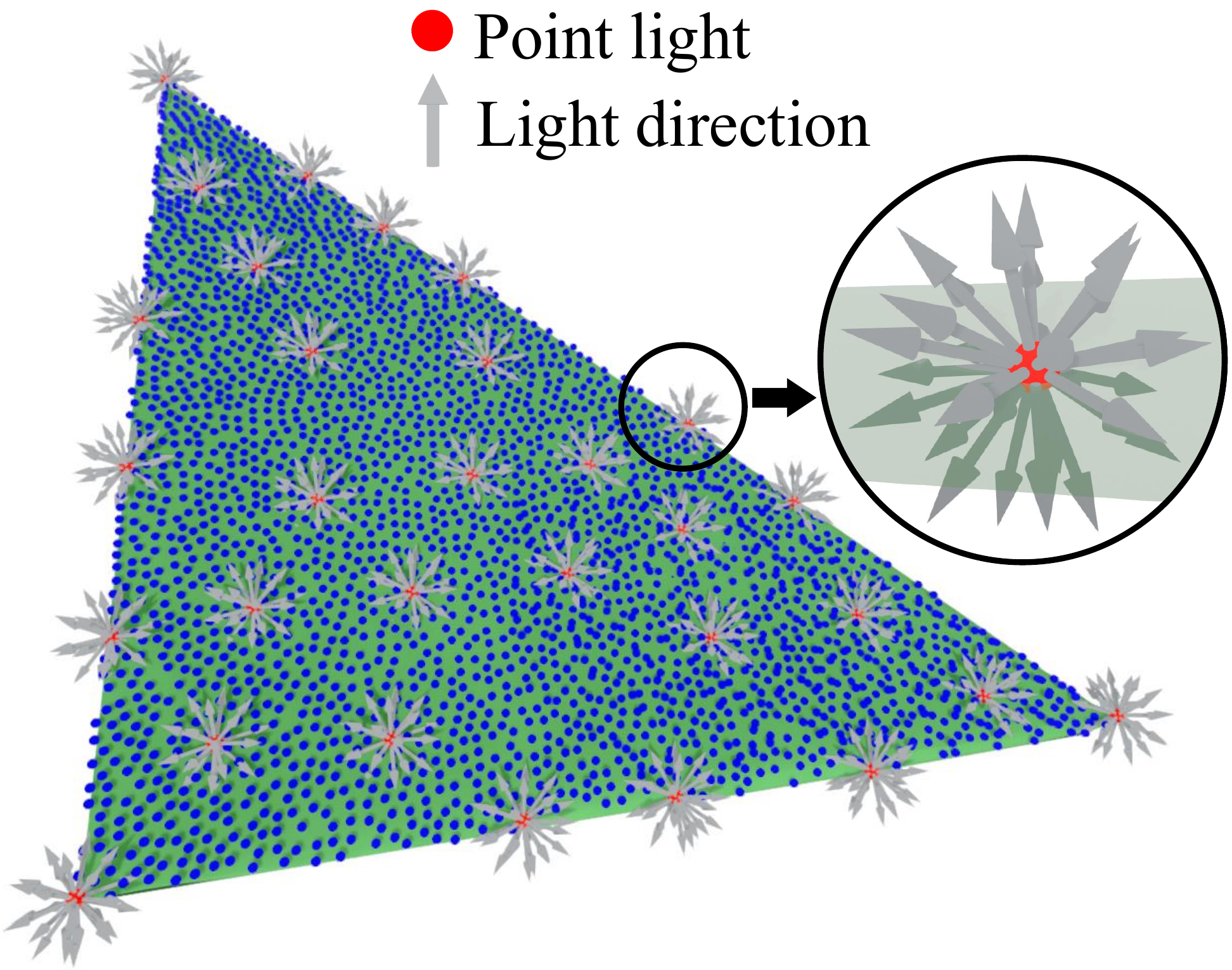} 
    \label{fig3}
\end{wrapfigure} 
farthest point sampling~\cite{qi2017pointnet++} on its supporting points to generate keypoints, as point light sources. Then uniform sphere sampling~\cite{gonzalez2010measurement} is used to generate lights (see inset). Thereafter, we compute the intersection of lights with all \( \alpha \)-shapes using the ray-shooting method~\cite{cgal:atw-aabb-24b}. If not intersect, this light is visible, otherwise is invisible. The visibility ratio of planes is obtained by dividing the number of visible light by the total number of samples. If the visibility ratio is larger than 0.5, the plane is defined as original highly visible, otherwise, barely visible. The detailed pseudo-code is provided in the supplementary material.

\begin{figure}[htbp]
  \centering
  \includegraphics[width=0.99999\linewidth]{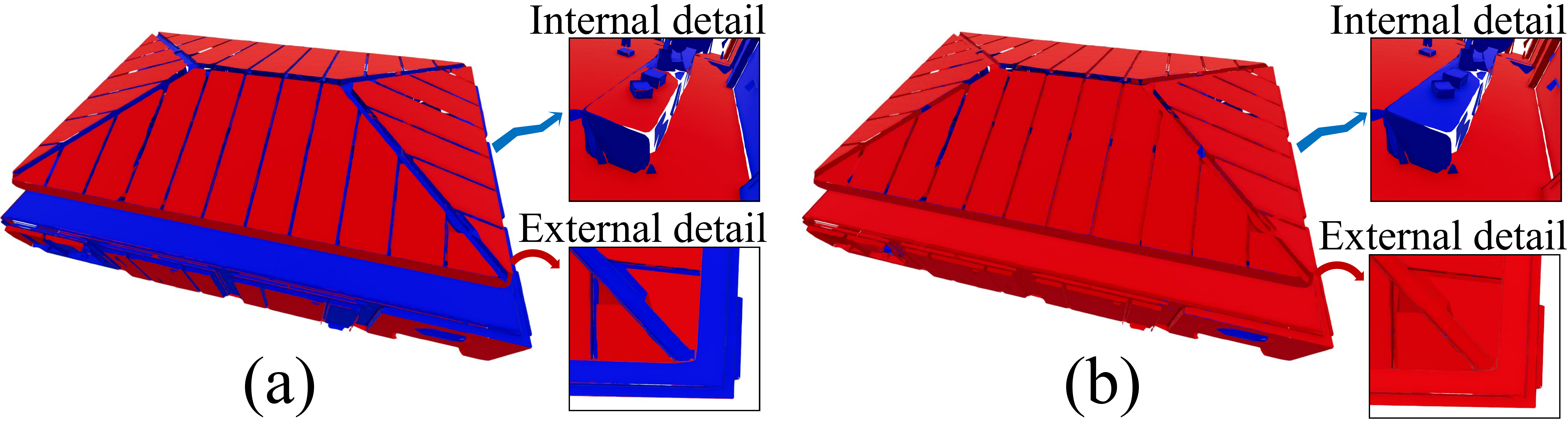}
  \caption{(a) Initial label assignment. (b) Optimized labels after energy minimization using MRF.}
  \label{fig:5}
\end{figure}

The visibility ratio of all planes is shown in Figure \ref{fig:2}(c1). It is basically consistent with spatial locations of planes, as is said, the closer a plane is to the external space, the greater the visibility ratio it got. However, some exceptions remain. For instance, the visibility ratio of the concave structure on the outside is less than 0.2 due to the obstruction of the surface patches in narrow spaces, as exemplified by the ceiling in Figure \ref{fig:5}(a). On the other hand, due to the holes caused by the missing scans, the visibility ratio of some inside planes is higher than 0.5, as shown by the table in Figure \ref{fig:5}(a). 
In order to improve this situation and make the label allocation of primitives more reasonable, we adopt an optimized scheme to assign visible labels to planes.

\textbf{Planar Labeling.}
We have the following observation: the neighbors of highly visible planes are also likely to be highly visible, and vice versa. Based on this neighborhood correlation, the MRF energy formulation is established. Specifically, given the set of planes \( \mathcal{P} \), the corresponding adjacency graph \( \mathcal{G} = (\mathcal{V}, \mathcal{E}) \), we aim to find a binary label \(x_i\), where \(x_i=1\) indicates highly visible and \(x_i=0\) indicates barely visible, to each primitive \(p_i\), whose visibility ratio is \(v_i\), using a MRF formulation:
\[E(x) = \sum_{p_i \in \mathcal{P}} E_{\text{data}}(p_i, x_i) + w \sum_{(i, j) \in \mathcal{E}} E_{\text{smooth}}(x_i, x_j)\]
The data term tends to assign \(x_i=1\) to planes with high visibility ratios:
\[E_{\text{data}}(p_i, x_i) = v_i \cdot (1 - x_i) + (0.5 - v_i) \cdot x_i,
\]while the smoothness term aims to ensure the consistency of neighboring labels:
\[E_{\text{smooth}}(x_i, x_j) = |x_i - x_j|.\]
where \( w \) is the balance parameters. In our experiments, we typically set \( w \) to 0.5. $E(x)$ is minimized with graph cuts and alpha expansion~\cite{boykov2004experimental}.

The final results are demonstrated in Figure \ref{fig:5}(b), where the visibility labels (highly/barely) of planes align consistently with both the visibility ratios and their spatial positions.

\subsection{Missing Plane Recovery}
A common characteristic of man-made objects is that each significant boundary line is typically formed by the intersection of two or more planes. Therefore, if a boundary segment is detected to be contained only by one single plane, then we consider this segment to be \textit{singular}. Using this as a clue, we first identify all the singular segments and then recover the missing planes based on them. 

\textbf{Singular segments selection.} For planar primitive \(p_i\), let \(S_i=\{s_{ij}\}\) be the set of boundary segments, and \(L_i=\{l_{ik}\}\) be the set of intersection lines between \(p_i\) and its second-order adjacent planar primitives. Two planes \(p_i\) and \(p_k\) are called second-order adjacent if there exists a third plane \(p_t\) that is adjacent to both \(p_i\) and \(p_k\). Segment \(s_{ij} \in S_i\) is considered not singular if there exists \(l_{ik} \in L_i\) such that: (i) The angle between the directions of \(l_{ik}\) and \(s_{ij}\) is less than \(r_a\), (ii) The distance from centroid of \(s_{ij}\) to \(l_{ik}\) is less than \(r_d\). In our approach, \( r_a \) is set to 10 degrees, and \( r_d \) is set to five times the average distance of the model. Furthermore, if \(s_{ij}\) is not singular, it should be contained by at least two planes, which in here are \(p_i\) and \(p_k\). 

\textbf{Missing planes generation.} 
A region-growing-liked method is used to fitting all planes from singular segments, as shown by the translucent planes in Figure \ref{fig:2}(d). Specifically, we sort segments based on their lengths in descending order and select seed segments accordingly, starting from the longest. Each seed segment determines an initial plane \({p_{ini}}\) based on its centroid and normal vector. Then a neighborhood search is performed around the seed segment, considering all segments within the second-order adjacent planes of the given segment's plane. A neighboring segment is incorporated into the region of the seed segment if two conditions are satisfied: (i) The angle between its normal vector and that of the initial plane \({p_{ini}}\) is less than a threshold \({r_a}\), (ii) The average distance of its supporting points to the initial plane \({p_{ini}}\) is below \({r_d}\). Then, the plane \({p_{ini}}\) is updated based on its region. This process is repeated until neighborhood searches for all segments have been completed. Planes containing at least two segments are selected as the final set of recovered planes. A detailed comparison between results with and without recovered planes is shown in Figure \ref{fig:6}. Comprehensive pseudo-code is in the supplementary material.

\begin{figure}[htbp]
  \centering
  \includegraphics[width=0.99999\linewidth]{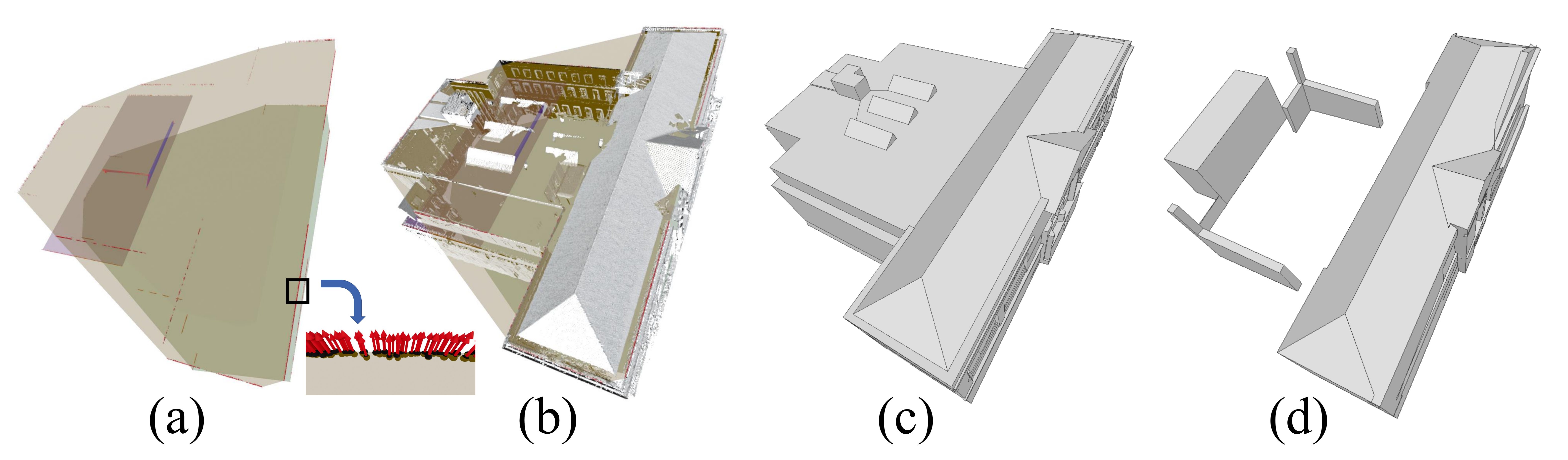}
  \caption{(a) the recovered missing planes, black points are support points of boundary segments, and the red arrows are their normal vectors reprojected from 2D into 3D. (b) recovered planes and point cloud model. (c) the output model with recovered planes. (d) the output model without recovered planes.}
  \label{fig:6}
\end{figure}

\subsection{Convex Hulls Generation}
\textbf{Visible planar convex hulls generation.} 
As said before, if a segment is not singular, it should be contained by the intersection line of two planes. Therefore, the supporting points of this segment will be incorporated into the calculation of the convex hulls in corresponding planes. 
In particular, for the primitive \(p_i\), we combine its original supporting points and the newly detected segments' supporting points to calculate the final expanded convex hull, as an effective compensation for missing data.

\textbf{Recovered planar convex hulls generation.} 
The computation of missing planar convex hulls is relatively straightforward, as it simply computes the convex hulls of corresponding boundary segments' supporting points.

\subsection{Hierarchical Space Partition}
\begin{figure}[htbp]
  \centering
  \includegraphics[width=0.99999\linewidth]{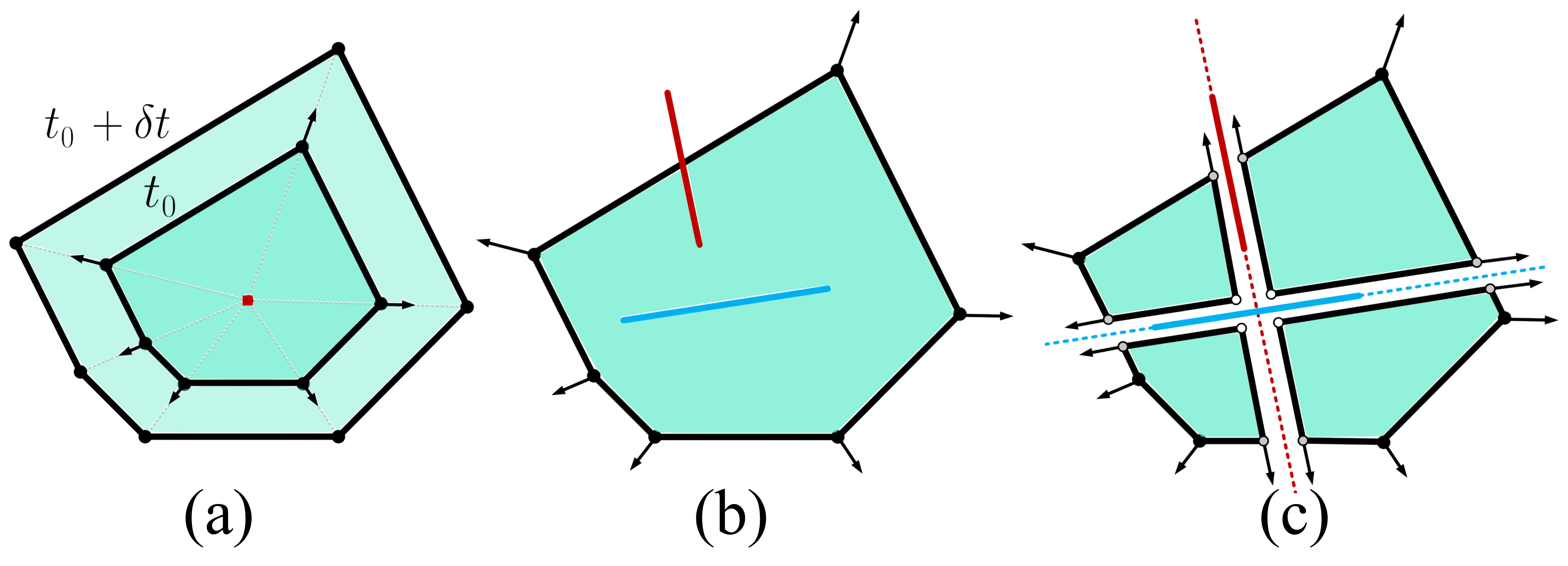}
  \caption{(a) The polygon grows by uniform scaling. (b) The intersection of the blue polygon with two others is illustrated by red and green line-segments. (c) The shape is decomposed into non-intersecting polygons by cutting along the intersection lines (shown as red and green dashed lines). Original vertices (black) retain their positions, while new vertices are either fixed at junctions (white, frozen vertices) or allowed to slide along intersection lines (gray, sliding vertices). The same as KSR~\cite{KSR:2020}.}
  \label{fig:8}
\end{figure}
\textbf{Background on kinetic framework}
The 3D kinetic framework originally developed by~\cite{KSR:2020}, which enables convex polygons to expand at consistent velocities (conformal transformation, see Figure \ref{fig:8}(a)), until collisions occur between each other, thereby partitioning the 3D space into polyhedral cells. This computational paradigm was first rigorously defined in the field of Computational Geometry by~\cite{guibas2018kinetic}. Within this framework, convex polygons have coordinates that evolve as continuous functions of time. The core mechanism involves maintaining spatial relationships among these primitives as they expand dynamically. 
When two or more primitives come into collisions (referred to as an \textit{event} in kinetic data structures), corrective geometric operations are executed to restore system validity, which is decomposing primitives into intersection-free polygons with cuts along the intersection lines, as show in Figure \ref{fig:8}(b-c). Algorithmically, the framework relies on a \textit{priority queue} to efficiently manage the temporal sequence of events, processing collisions iteratively until all primitives reach stationary states~\cite{KSR:2020}.

\textbf{Hierarchical strategy.} 
Based on kinetic framework, a hierarchical space partition strategy is introduced. This strategy shares the same objective as other partitioning strategies: minimizing unnecessary splits~\cite{chauve2010robust, sulzer2024concise, pan2025building}. However, unlike approaches that prioritize planes based on size or other attributes, we employ visibility ratios to categorize primitives into distinct levels (i.e., level 1: highly visible planes; level 2: barely visible planes; level 3: invisible planes (missing planes)). Then, the lower the level, the first to grow, and the space is partitioned progressively. This insight is based on our observation that primitives exhibiting higher visibility ratios—typically located near the periphery—should be prioritized and expanded faster during the growth process, thereby forming boundaries for inner planes with lower visibility ratio. 
\begin{figure}[htbp]
  \centering
  \includegraphics[width=0.98\linewidth]{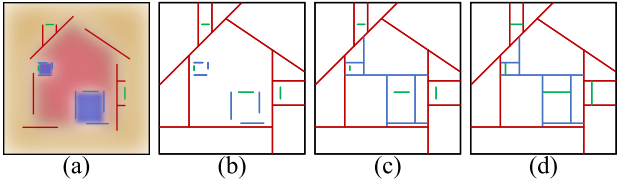}
  \caption{Different colors represent different visible hierarchical regions (a), red line-segments (i.e., highly visible planes) grows first (b), then blue grows (barely visible planes) (c), and finally green grows (invisible planes) (d).}
  \label{fig:101}
\end{figure}

Specifically, as shown in Figure \ref{fig:101}, let the bounding box be the level 0, and the corresponding space be space-0. The highly visible planes first perform space partition based on space-0 to generate the space-1. Barely visible planes then grow to generate space-2 based on space-1. And finally the missing planes expand to subpart the space-2 to generate space-3. Namely, the (i+1)-th space is generated by planes of (i+1) level to subpart the (i)-th space (i = 0, 1, 2). The facets generated in the previous level are equivalent to the bounding box of current level. And the space-3 is what we sought. Also as shown in Figure \ref{fig:2}(f). See supplementary material for detailed pseudo-code. Moreover, unlike~\cite{KSR:2020, PolyFit}, which handle all planes uniformly, we introduce a unique expand speed for each primitive and a more natural condition for convex hulls to stop growing.

\textit{Expanding speed}. The growth of each plane gives a speed positively correlated with the visibility ratio; that is, the faster the primitive with a higher visible rate will grow in the same level. Specifically, for each level, the visibility ratios of its planes are normalized to a unified range of [0.5, 1], where the minimum value corresponds to 0.5 and the maximum value corresponds to 1, and this defines their growth speed.
\begin{figure}[htbp]
  \centering
  \includegraphics[width=0.99999\linewidth]{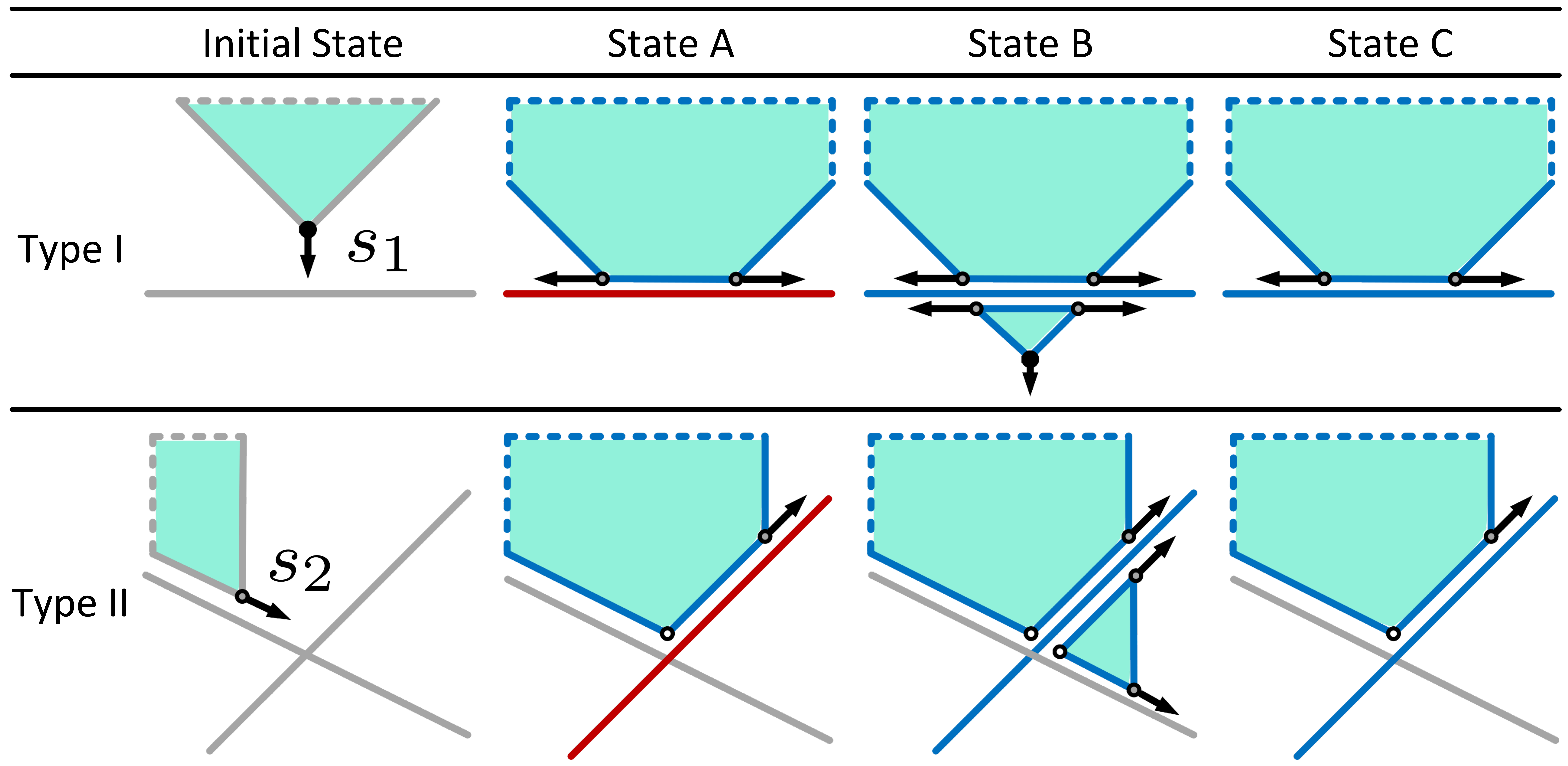}
  \caption{Different polygon collision types and growth states. $s_1$ and $s_2$ represent different speeds. The red lines indicate highly visible polygons, the blue lines indicate barely visible polygons, and the gray lines denote polygons that may be any type. The point colors follow the same definition as in Figure~\ref{fig:8}.}
  \label{fig:9}
\end{figure}

\textit{State Transition}. 
We treat the vertices of each polygon as a state machine to update the growth process. The initial state is when all the polygons are divided into non-intersecting sub-polygons before they begin to grow, as shown in Figure \ref{fig:8}(c). Different state transitions are triggered based on vertices and different types of plane encounters (\textit{events}): if a vertex meets the plane of the previous level (\textit{events A}) or the adjacent plane of the current level (\textit{events C}), the direction of propagation of it will be modified to follow the intersection line with the intersecting plane. 
Specifically, for states A and C in type I in Figure \ref{fig:9}, we split the vertex into two sliding vertices diverging along the intersection line. For type II states A and C, we reorient the sliding vertex propagation to align with the target polygon's intersection. Meanwhile, a frozen vertex is established at the convergence point. In addition to the above events, collisions also occur between non-neighboring primitives at the same level (\textit{events B}). In this case, it is necessary to introduce a newly generated polygon into the kinetic data structure. This polygon extends the originating primitive across the shared boundary defined by the intersecting polygon (see state B in Figure \ref{fig:9}). The initialization of this polygon differs depending on the configuration: in state B of type I, it consists of two sliding vertices and one original vertex; whereas in state B of type II, it is composed of two sliding vertices along with one fixed vertex. 
Finally, we present the terminal state: a sliding vertex reaches the intersection line, at a moment when another sliding vertex—guided by the contacted polygon—is already in place. At the intersection point of the corresponding lines, a frozen vertex is introduced, effectively halting the local propagation of both sliding vertices.

Accordingly, we use the boundary and neighborhood relationship to constrain growth. Thanks to the expansion of visible primitives and the addition of missing planes, which ensures the effective completion of the missing data, so only a few growth events are needed to achieve an effective spatial division, ensuring detail recovery, completing missing structures, and reducing unnecessary partitions.

\subsection{Surface Extraction}
We extract the surface by performing a min-cut on the polyhedral partition, assigning inside-outside labels to the 
\setlength{\intextsep}{2pt}
\setlength{\columnsep}{4pt}
\begin{wrapfigure}{r}{0.25\textwidth} 
    \centering
    \includegraphics[width=0.9\linewidth]{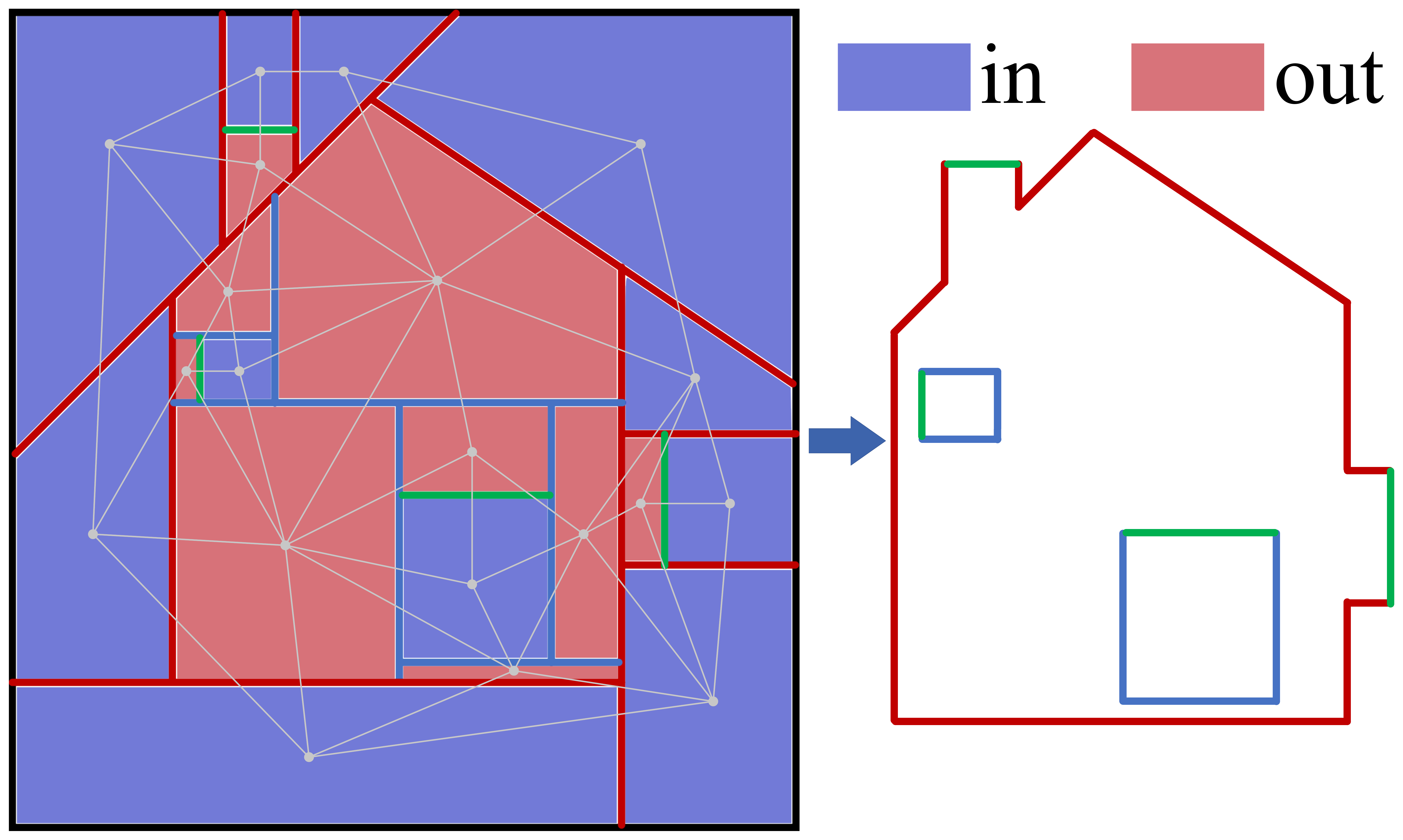} 
    \label{fig:17}
\end{wrapfigure}
cells. The resulting surface corresponds to the interface facets between labeled regions (see inset). Owing to the validity of the kinetic embedding, the output is guaranteed to be watertight and free of self-intersections, similar to \cite{KSR:2020, chauve2010robust, xiang2024efficient, li2016manhattan}.

Given a partitioned polyhedral set $\mathcal{C} = \{c_i\}$, we construct its dual graph $\mathcal{G} = (\mathcal{V}, \mathcal{E})$ as follows: (i) For each polyhedron $c_i \in \mathcal{C}$, create a vertex $v_i \in \mathcal{V}$ positioned at $c_i$'s centroid. (ii) For each pair $(c_i, c_j)$ sharing a facet $f_{ij}$, add an edge $e_{ij} \in \mathcal{E}$ connecting $v_i$ and $v_j$, as shown in inset.

To assign binary labels \( x_i \in \{0, 1\} \) to each node \( v_i \), where \( x_i = 0 \) denotes that \( c_i \) lies outside and \( x_i = 1 \) denotes that it lies inside, we define the following energy function \( U \):
\[U(\mathbf{x}) = P(\mathbf{x}) + \lambda M(\mathbf{x}) \tag{2}\]
where $\lambda \in [0, 1]$ is a balance parameter, which is set to 0.5 in our methods. 
The optimal output surface that minimizes $U$ is found by a max-flow algorithm \cite{boykov2004experimental}.

Following the convention used in smooth surface reconstruction \cite{kazhdan2013screened}, normals are assumed to point outward. Accordingly, the point supporting term $P(\mathbf{x})$ evaluates the consistency between the assigned labels and the orientations of inlier normals:
\[
P(\mathbf{x}) = \frac{\pi \cdot {r}^2}{\mathcal{A}} \sum_{v_{i} \in \mathcal{V}} \sum_{e_{ij} \in \mathcal{E}} \sum_{p_{k} \in I_{ij}} \mathbf{1}_{\{(2x_i - 1) \cdot \vec{n}_k \cdot \vec{u}_{ij}<0\}} 
\tag{3}
\]
\setlength{\intextsep}{2pt}
\setlength{\columnsep}{4pt}
\begin{wrapfigure}{r}{0.17\textwidth} 
    \centering
    \includegraphics[width=0.9\linewidth]{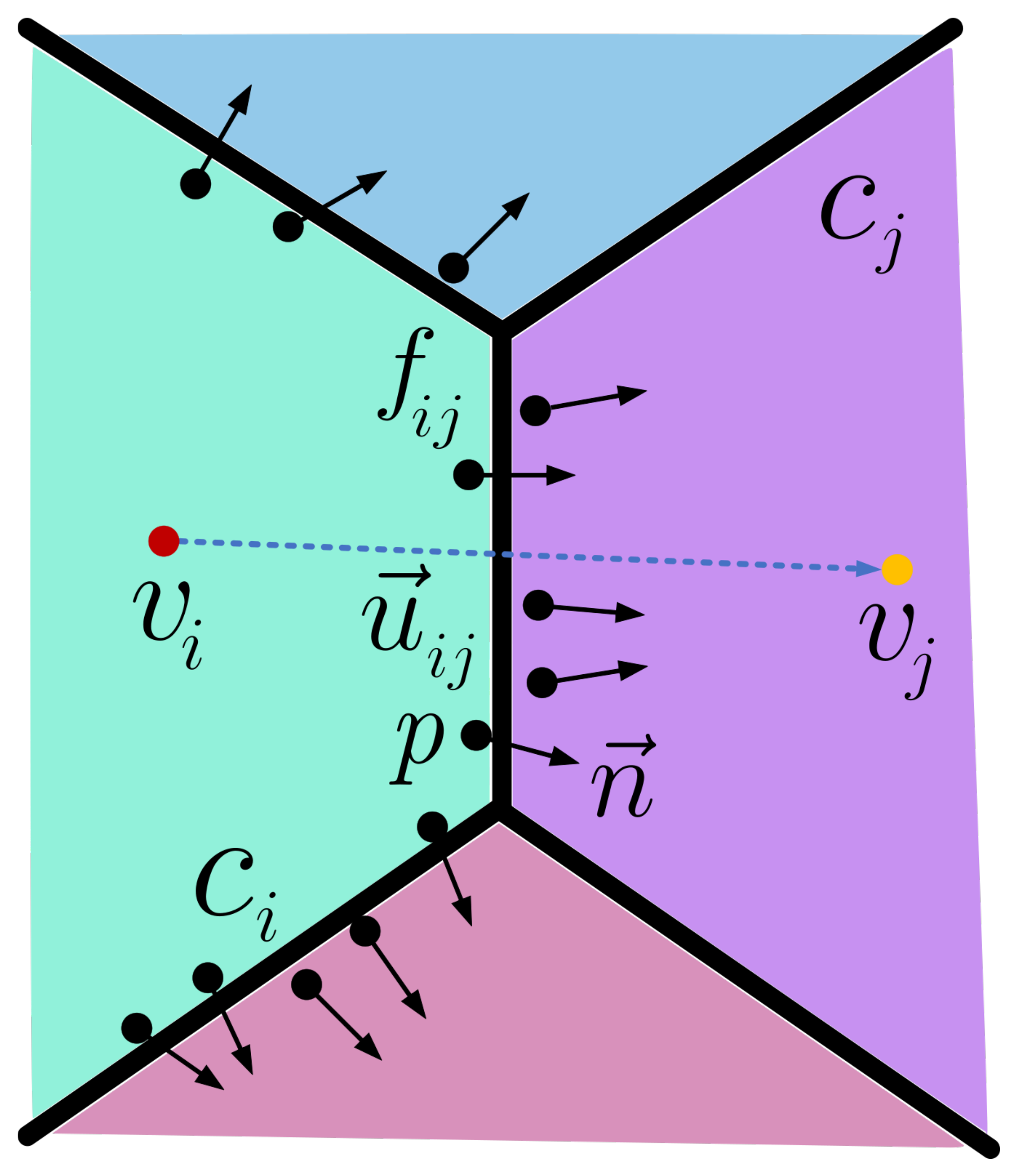} 
    \label{fig:187}
\end{wrapfigure} 
where $\mathcal{A}$ is a normalization factor defined as the sum of the areas of all facets of the partition, $I_{ij}$ is the set of inlier points associated with common facet $f_{ij}$ between $c_i$ and $c_j$. $\mathbf{1}_{\{ \circ \}}$ is the characteristic function, $\vec{n}_k$ is the normal vector of inlier point $p_k$, and $\vec{u}_{ij}$ is the vector from $v_i$ to $v_j$, as shown in the inset.
${r}$ is the average spacing of input point cloud model. In this example, we prefer assigning label \textit{in} to polyhedron $c_i$ (i.e., $x_i$=1) and label \textit{out} to polyhedron $c_j$ (i.e., $x_j$=0). 
We use the centroid line $\vec{u}_{ij}$ of the two polyhedra $c_i$ and $c_j$ to determine the points on their common surface $f_{ij}$. This local neighborhood context makes our voting function more robust to inaccuracies in normal orientations, since normals only need to point toward the correct half-space separating the facet. 

The regularization term $M(\mathbf{x})$ penalizes surface complexity through area minimization, where reduced area correlates with simplified geometry. Furthermore, facets with abundant supporting points exhibit higher retention likelihood during optimization:
\[
M(\mathbf{x}) = \frac{1}{\mathcal{A}} \sum_{e_{ij} \in \mathcal{E}} (a_{ij} - \pi \cdot r^2 \cdot |I_{ij}| ) \cdot \mathbf{1}_{\{x_i \ne x_j\}} \tag{4}
\]
where $a_{ij}$ represents the area of facet $f_{ij}$, and $|I_{ij}|$ is the number of inlier points associated with facets $f_{ij}$.

\section{Results}
\label{sec:result}
\begin{figure*}[t]
  \centering
  \includegraphics[width=0.99\textwidth]{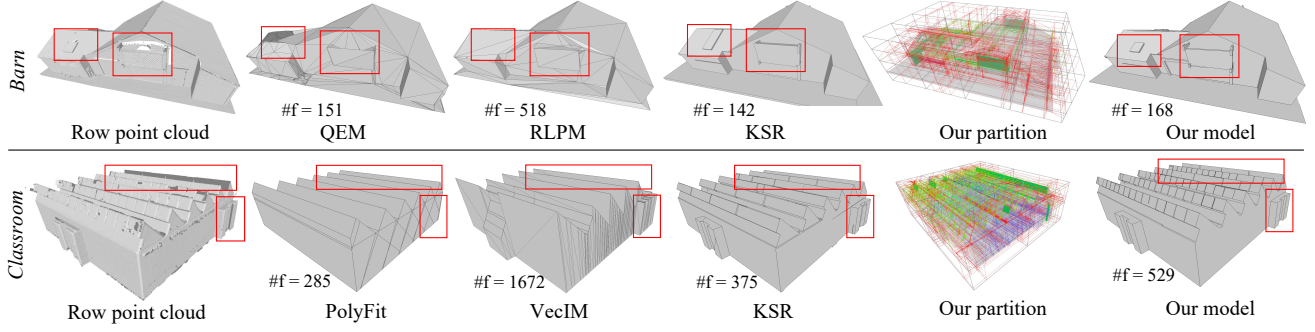}
  \caption{Reconstruction results from different methods on the urban scene \textit{Barn} and the indoor scene \textit{Classroom}. \#f refers to the number of output facets. Our method achieves the most accurate reconstructions, with local details highlighted in red boxes. In the subfigures labeled "Our partition", green planes indicate the recovered planar regions.}
  \label{fig:11}
\end{figure*}
Our algorithm is implemented in C++ using the CGAL~\cite{cgal:eb-24b} library. 
We evaluate the effectiveness of our method for mesh reconstruction under a standard experimental setup with an Intel i5-10300H CPU and an NVIDIA GTX 1650 GPU. 
The evaluation includes benchmark datasets, quantitative metrics, and both objective and subjective comparisons with contemporary reconstruction methods. 
Due to space constraints, parameter sensitivity analysis, ablation studies, and generalization experiments are provided in the supplementary material.

\textbf{Dataset and Metrics.} 
For benchmarking purposes, test point clouds from the Assembly Dataset~\cite{willis2022joinable}, the KSR-42 Dataset~\cite{KSR:2020}, and the ScanNet++ V2 Dataset~\cite{yeshwanth2023scannet++} are employed, as they exhibit varying degrees of reconstructed geometric fidelity. Inspired by~\cite{he2024windpoly} and guided by the data scale in~\cite{gao2022low}, we select 200 CAD models (\textit{CAD\text{-}200}) and 100 real scanned architectural models (\textit{Arch\text{-}100}) from these datasets—that exhibit diverse geometric characteristics, to evaluate the reconstruction performance. We assess our method in terms of correctness, conciseness, and computational performance. Correctness is quantified using Mean Hausdorff Error (\textit{MHE}) and Root Mean Squared Error (\textit{RMSE}), reflecting how well the reconstructed polygonal planes align with the input points. Conciseness is evaluated based on the number of vertices ($\text{P}^{\text{Avg.}}$) and facets ($\text{F}^{\text{Avg.}}$) in the reconstructed mesh, as well as a composite metric ($\text{RH}^{\text{Avg.}}$) that combines the Hausdorff distance and a simplification ratio. The simplification ratio is calculated as the number of vertices in the reconstructed mesh divided by the number of points in the input point cloud, following the definition in~\cite{he2024windpoly}. Computational efficiency is assessed via runtime and peak memory usage during reconstruction.

\textbf{Comparative analysis.}  
Based on the above dataset and evaluation criteria, we compare our method against several representative approaches, including QEM~\cite{garland1997surface}, PolyFit~\cite{PolyFit}, KSR~\cite{KSR:2020}, Robust Low-Poly Meshing (RLPM)~\cite{chen2023robust}, and VecIM~\cite{han2021vectorized}, which reflect current mainstream strategies. Since QEM and RLPM are incapable of directly processing raw point clouds, we first apply the Poisson surface reconstruction technique~\cite{kazhdan2013screened} to generate triangular meshes for these methods. VecIM primarily focuses on indoor scene reconstruction and requires accurate semantic information for guidance, so we only test this method on indoor scenes. Excessive primitives make PolyFit's mixed-integer optimization costly; thus, keeping their number below 100 ensures feasible vectorization, though some local details may be lost.
\begin{table}[!htbp]
\centering
\renewcommand{\arraystretch}{0.95}
\resizebox{\linewidth}{!}{
\begin{tabular}{l@{\hskip 5pt}c@{\hskip 5pt}c@{\hskip 9pt}c@{\hskip 9pt}c@{\hskip 9pt}c@{\hskip 9pt}c@{\hskip 5pt}c@{\hskip 5pt}c}
\toprule
& & \multicolumn{2}{c}{\textit{Correctness}} & \multicolumn{3}{c}{\textit{Conciseness}} & \multicolumn{1}{c}{\textit{Time}} \\
\cmidrule(lr){3-4} \cmidrule(lr){5-7} \cmidrule(lr){8-8}
& & $\text{MHE} \downarrow$ & $\text{RMSE} \downarrow$ & $\text{P}^{\text{Avg.}} \downarrow$ & $\text{F}^{\text{Avg.}} \downarrow$ & $\text{RH}^{\text{Avg.}} \downarrow$ & T(s) $\downarrow$ \\
\midrule
\multirow{6}{*}{\rotatebox{90}{$Arch\text{-}100$}}
& QEM~\cite{garland1997surface}      &  2.702 & 0.082 & 602 & 811 & 0.0137 & \textbf{76} \\
& Polyfit~\cite{PolyFit}             &  2.165 & 0.059 & 516 & 723 & 0.0061 & 1253 \\
& KSR~\cite{KSR:2020}  &  1.926 & 0.037 & \textbf{260} & \textbf{372} & 0.0057 & 265 \\
& VecIM~\cite{han2021vectorized}     &  2.962 & 0.096 & 712 & 1215 & 0.0129 & 152 \\
& RLPM~\cite{chen2023robust}         &  2.298 & 0.078 & 565 & 761 & 0.0102 & 351 \\
& Ours              & \textbf{1.066} & \textbf{0.025} & 315 & 501 & \textbf{0.0036} & 327 \\
\midrule
\multirow{4}{*}{\rotatebox{90}{$CAD\text{-}200$}}
& PolyFit~\cite{PolyFit}        & 0.159    & 0.015 & 371          & 559          & 0.0023 & 867 \\
& KSR~\cite{KSR:2020}           & 0.127    & 0.016 & \textbf{292} & \textbf{314} & 0.0015 & \textbf{182} \\
& RLPM~\cite{chen2023robust}    & 0.236    & 0.037 & 421          & 926          & 0.0025 & 291 \\
& Ours           & \textbf{0.109} & \textbf{0.009} & 265          & 331  & \textbf{0.0012} & 205 \\
\bottomrule
\end{tabular}
}
\caption{Comparison of correctness, conciseness, and runtime.}
\label{tab:comparison}
\end{table}

\begin{figure}[htbp]
  \centering
  \includegraphics[width=0.99\linewidth]{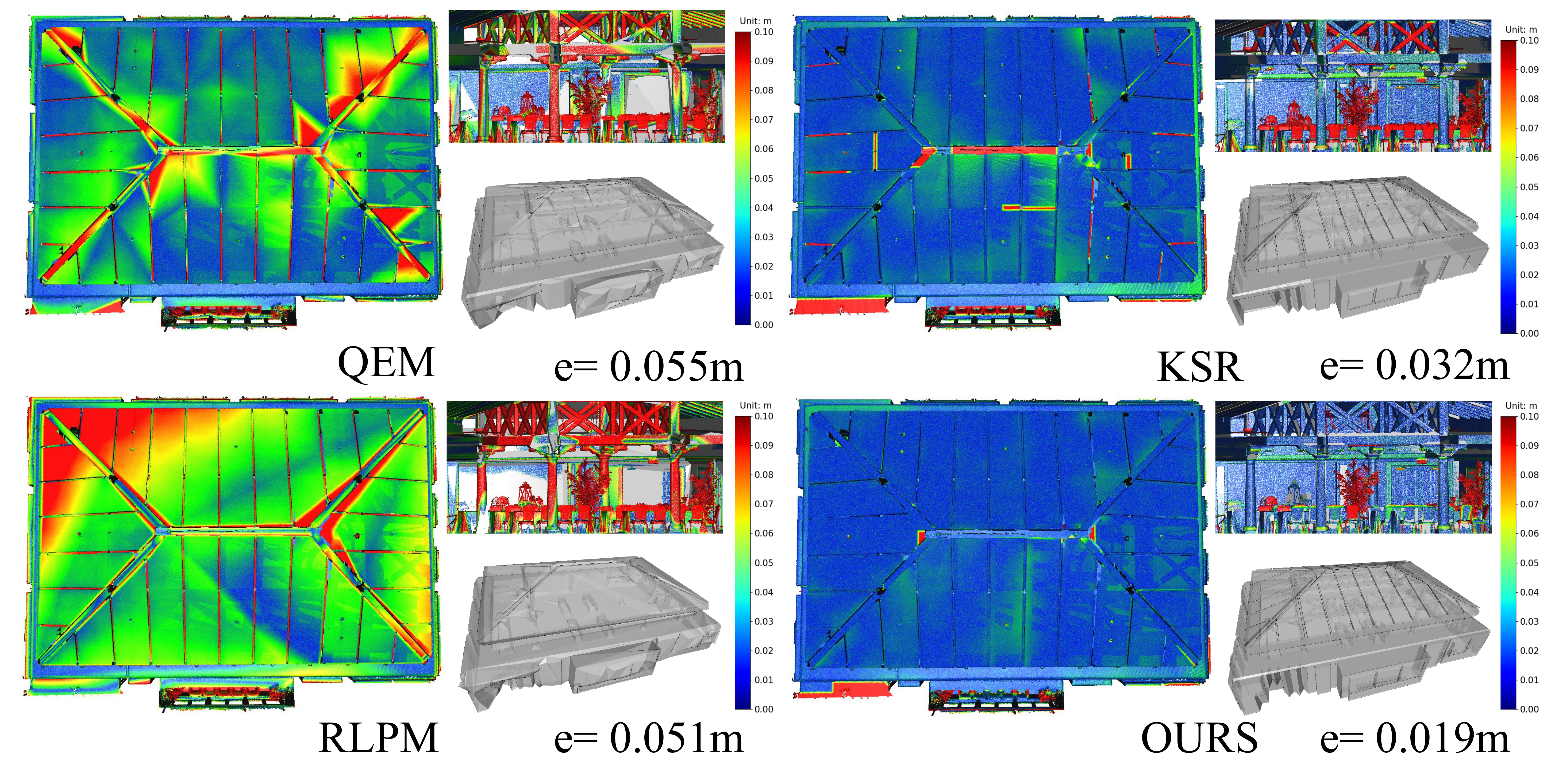}
  \caption{Output models and distance maps from various methods on the Meetingroom. e: RMSE from input points to output model.}
  \label{fig:15-1}
\end{figure}

\textit{Correctness.} 
As representative examples, the reconstruction results on the \textit{Barn} and \textit{Classroom} (C.Room) scenes are shown in Figure \ref{fig:11}. As shown in the figures, our algorithm's missing planes recovery technique enables the faithful reconstruction of fine details, such as the window sill and skylights in the \textit{Barn} scene. Similarly, the ceiling and the lower wall in the \textit{Classroom} scene are also accurately recovered. The green planes in Figure \ref{fig:11} indicate the recovered planes, corresponding to regions that are missing or unscanned during the LiDAR acquisition. These regions significantly affect the overall reconstruction accuracy. While planar expansion helps compensate for some of these deficiencies, it remains an imperfect solution~\cite{KSR:2020, PolyFit}. A quantitative comparison of reconstruction accuracy is presented in Figure \ref{fig:15-1} and Table~\ref{tab:comparison}. Figure \ref{fig:15-1} shows that our algorithm successfully reconstructs the internal beam structure of the \textit{Meetingroom} (M.Room), which is missing in the point cloud due to occlusion, leading to a significantly reduced reconstruction error. Table~\ref{tab:comparison} confirms that our method attains superior reconstruction accuracy compared to other algorithms.
\begin{figure*}[t]
  \centering
  \includegraphics[width=0.99\linewidth]{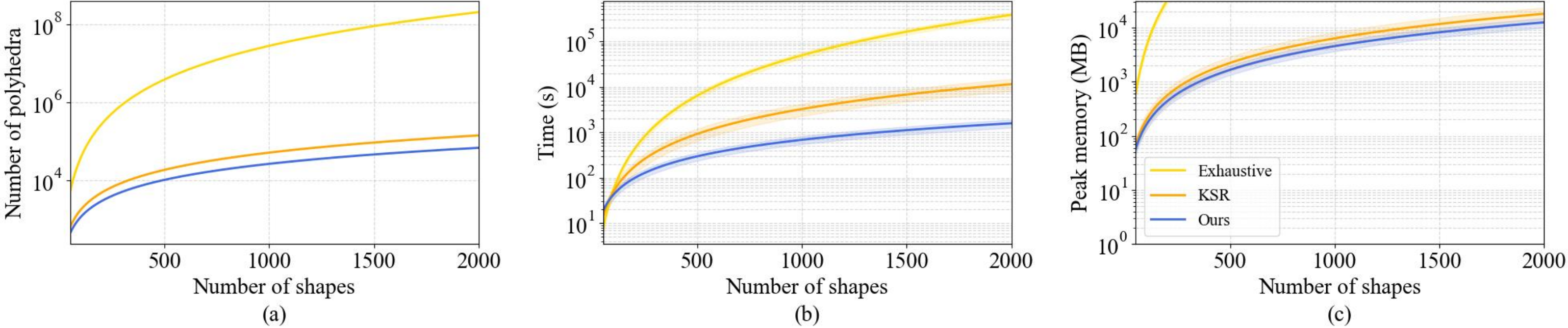}
  \caption{Performances of partitioning and surface extraction. The transparent band around each curve indicates the minimal and maximal values measured on various models.}
  \label{fig:12}
\end{figure*}

\begin{figure}[!htbp]
  \centering
  \includegraphics[width=0.99\linewidth]{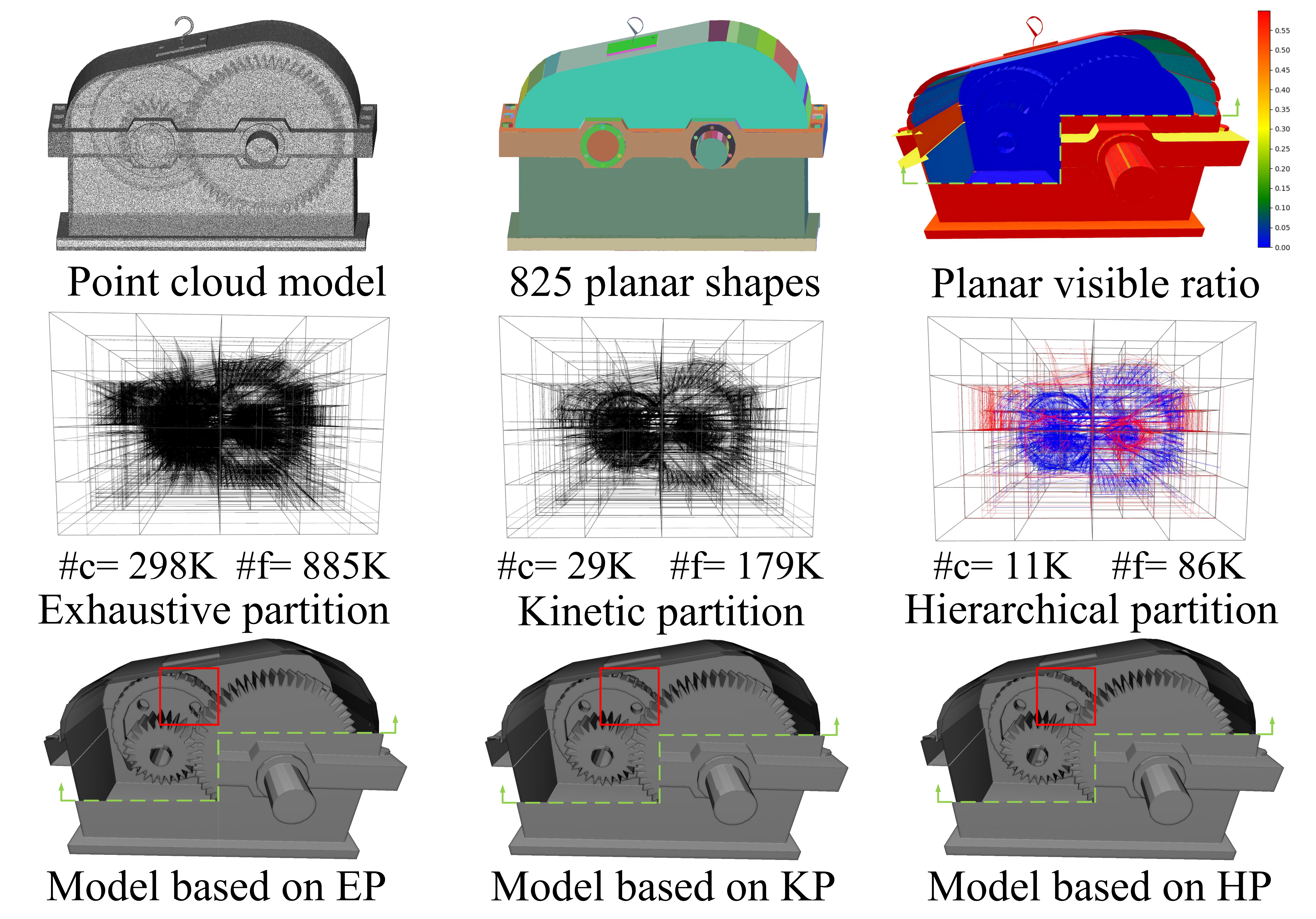}
  \caption{Exhaustive, kinetic, and hierarchical partitions on the $Gearbox$. 
  \#f and \#c denote the numbers of facets and polyhedra, respectively. 
  Our hierarchical partitions (HP) are more compact and yield more accurate reconstructions than exhaustive (EP) and kinetic partitions (KP), as highlighted in red. 
  The model is sliced along the green dashed line for internal visualization.}
  \label{fig:15}
\end{figure}

\textit{Conciseness.}
In terms of conciseness, as shown in Table~\ref{tab:comparison}, our method reconstructs the model with slightly more vertices and facets due to an increased number of planes (i.e., the recovered planes). However, it achieves the highest overall compactness and the best trade-off between facets count and reconstruction accuracy. Further, in order to prove the superiority of our hierarchical spatial partition, we made a more detailed comparison on the \textit{CAD\text{-}200}. Since the \textit{CAD\text{-}200} point cloud is sampled from complete CAD meshes, no planes are missing, so all plane assembly schemes yield the same number of primitives. In Figures \ref{fig:12} and \ref{fig:15}, it can be seen that our algorithm achieves the most compact spatial division, owing to a hierarchical strategy and the scheme of correlating the growth speed of convex hulls with the visibility rate that avoids meaningless partitions. Moreover, by expanding plane polygons along contained line-segments (Section~\ref{sec:method}) and stopping growth based on neighborhood cues, our method enables each plane to partition the required polyhedral cells with minimal extension, achieving both compactness and accuracy. As shown in Figure \ref{fig:12}(a), our method produces the fewest polyhedral cells and demonstrates the slowest growth in cell count across different planar motifs.

\textit{Efficiency.}
Our method is slightly slower than KSR but significantly faster than PolyFit, as shown in Table~\ref{tab:comparison}. 
For the plane classification module, thanks to our use of a bounding volume hierarchy~\cite{cgal:atw-aabb-24b} and a fixed number of sampled keypoints per plane, the overall algorithm complexity is slightly above $\text{O}(n)$ but significantly lower than $\text{O}(n^2)$, where $n$ is the number of planes. 
The complexity of the missing plane recovery module is similar to that of the plane classification module, since only a few number of segments ultimately pass the screening. The most time-consuming part of our algorithm is the hierarchical spatial partitioning module, as shown in the runtime breakdown in Table~\ref{tab:2}. Moreover, due to our neighborhood constraint strategy and planar hierarchical growth strategy, our memory peak and partition time are the lowest, as shown in Figure \ref{fig:12}.

\begin{table}[!htbp]
  \centering
  \small 
  \setlength{\tabcolsep}{3pt}
  \begin{tabular}{l c c c c c}
    \toprule
                             & M.Room & Barn & C.Room & Gearbox \\
    \midrule
    \# input points          & 3.07M & 2.01M & 3.98M & 1.85M  \\
    \# highly visible planes & 872   & 157   & 175   & 209  \\
    \# barely visible planes & 815   & 38    & 80    & 616  \\
    \# recovered planes      & 35    & 19    & 38    & no  \\
    \# output facets         & 1929  & 168   & 529   & 2010  \\
    Categorization (sec)     & 91    & 12    & 21    & 76  \\
    Recovery (sec)           & 79    & 10    & 35    & no  \\
    Partitioning (sec)       & 219   & 67    & 123   & 192  \\
    Extraction (sec)         & 71    & 25    & 57    & 75  \\
    Memory peak (MB)         & 675   & 152   & 335   & 596  \\
    \bottomrule
  \end{tabular}
  \caption{Performance statistics of different scenes.}
  \label{tab:2}
\end{table}

\section{Conclusion}
\label{sec:conclusion}
We propose a novel algorithm that reconstructs compact and watertight polygonal meshes from raw point clouds.
The main contribution of this work is the introduction of hierarchical space partitioning strategy, where the space are progressively partitioned by prioritizing the growth of planes according to their visibility ratio into a low number of polyhedra. This idea not only improves correctness over existing methods by recovering missing scene details, but also enables the generation of more concise polygonal meshes from a more compact and semantically meaningful partition of polyhedra.
We conducted extensive comparative experiments on diverse datasets against state-of-the-art methods, demonstrating the accuracy and effectiveness of our approach. In future work, we intend to incorporate additional available information, such as multi-view stereo image data, to capture richer spatial structures.
We also plan to investigate how the proposed hierarchical space partition strategy can be applied to the repair of CAD models.

\noindent\textbf{Acknowledgments.}
This work was supported by the National Natural Science Foundation of China under Grant Nos. 52575574, U24A20130 and 52188102.

{
    \small
    \bibliographystyle{ieeenat_fullname}
    \bibliography{main}

@String(ICCV= {Int. Conf. Comput. Vis.})

@String(TOG= {ACM Trans. Graph.})

@String(ICCV  = {ICCV})

@String(TOG   = {ACM TOG})

@article{KSR:2020,
author = {Bauchet, Jean-Philippe and Lafarge, Florent},
title = {Kinetic Shape Reconstruction},
year = {2020},
issue_date = {October 2020},
publisher = {Association for Computing Machinery},
address = {New York, NY, USA},
volume = {39},
number = {5},
issn = {0730-0301},
url = {https://doi.org/10.1145/3376918},
doi = {10.1145/3376918},
journal = {ACM Trans. Graph.},
month = jun,
articleno = {156},
numpages = {14}
}

@article{PolyFit,
author={Nan, Liangliang and Wonka, Peter},
booktitle={2017 IEEE International Conference on Computer Vision (ICCV)}, 
title={PolyFit: Polygonal Surface Reconstruction from Point Clouds}, 
year={2017},
volume={},
number={},
pages={2372-2380},
doi={10.1109/ICCV.2017.258}}

@article{fang2021structure,
  title={Structure-aware indoor scene reconstruction via two levels of abstraction},
  author={Fang, Hao and Pan, Cihui and Huang, Hui},
  journal={ISPRS Journal of Photogrammetry and Remote Sensing},
  volume={178},
  pages={155--170},
  year={2021},
  publisher={Elsevier}
}

@inproceedings{sulzer2024concise,
  title={Concise Plane Arrangements for Low-Poly Surface and Volume Modelling},
  author={Sulzer, Raphael and Lafarge, Florent},
  booktitle={European Conference on Computer Vision},
  pages={357--373},
  year={2024},
  organization={Springer}
}

@inproceedings{chauve2010robust,
  title={Robust piecewise-planar 3D reconstruction and completion from large-scale unstructured point data},
  author={Chauve, Anne-Laure and Labatut, Patrick and Pons, Jean-Philippe},
  booktitle={2010 IEEE computer society conference on computer vision and pattern recognition},
  pages={1261--1268},
  year={2010},
  organization={IEEE}
}

@incollection{guibas2018kinetic,
  title={Kinetic data structures},
  author={Guibas, Leonidas},
  booktitle={Handbook of Data Structures and Applications},
  pages={377--388},
  year={2018},
  publisher={Chapman and Hall/CRC}
}

@article{han2021vectorized,
  title={Vectorized indoor surface reconstruction from 3D point cloud with multistep 2D optimization},
  author={Han, Jiali and Rong, Mengqi and Jiang, Hanqing and Liu, Hongmin and Shen, Shuhan},
  journal={ISPRS Journal of Photogrammetry and Remote Sensing},
  volume={177},
  pages={57--74},
  year={2021},
  publisher={Elsevier}
}

@article{xiang2024efficient,
  title={Efficient High-Quality Vectorized Modeling of Large-Scale Scenes},
  author={Xiang, Xiaojun and Jiang, Hanqing and Yu, Yihao and Shen, Donghui and Zhen, Jianan and Bao, Hujun and Zhou, Xiaowei and Zhang, Guofeng},
  journal={International Journal of Computer Vision},
  volume={132},
  number={10},
  pages={4564--4588},
  year={2024},
  publisher={Springer}
}

@article{huang2023arrangementnet,
  title={ArrangementNet: learning scene arrangements for vectorized indoor scene modeling},
  author={Huang, Jingwei and Zhang, Shanshan and Duan, Bo and Zhang, Yanfeng and Guo, Xiaoyang and Sun, Mingwei and Yi, Li},
  journal={ACM Transactions on Graphics (TOG)},
  volume={42},
  number={4},
  pages={1--15},
  year={2023},
  publisher={ACM New York, NY, USA}
}

@article{chen2022,
  title={Reconstructing compact building models from point clouds using deep implicit fields},
  author={Chen, Zhaiyu and Ledoux, Hugo and Khademi, Seyran and Nan, Liangliang},
  journal={ISPRS Journal of Photogrammetry and Remote Sensing},
  volume={194},
  pages={58--73},
  year={2022},
  publisher={Elsevier}
}

@inproceedings{fang2020connect,
  title={Connect-and-slice: an hybrid approach for reconstructing 3d objects},
  author={Fang, Hao and Lafarge, Florent},
  booktitle={Proceedings of the IEEE/CVF Conference on Computer Vision and Pattern Recognition},
  pages={13490--13498},
  year={2020}
}

@inproceedings{boulch2014piecewise,
  title={Piecewise-planar 3D reconstruction with edge and corner regularization},
  author={Boulch, Alexandre and de La Gorce, Martin and Marlet, Renaud},
  booktitle={Computer Graphics Forum},
  volume={33},
  number={5},
  pages={55--64},
  year={2014},
  organization={Wiley Online Library}
}

@inproceedings{langlois2019surface,
  title={Surface reconstruction from 3d line segments},
  author={Langlois, Pierre-Alain and Boulch, Alexandre and Marlet, Renaud},
  booktitle={2019 International Conference on 3D Vision (3DV)},
  pages={553--563},
  year={2019},
  organization={IEEE}
}

@inproceedings{li2016manhattan,
  title={Manhattan-world urban reconstruction from point clouds},
  author={Li, Minglei and Wonka, Peter and Nan, Liangliang},
  booktitle={Computer Vision--ECCV 2016: 14th European Conference, Amsterdam, The Netherlands, October 11--14, 2016, Proceedings, Part IV 14},
  pages={54--69},
  year={2016},
  organization={Springer}
}

@article{oesau2014indoor,
  title={Indoor scene reconstruction using feature sensitive primitive extraction and graph-cut},
  author={Oesau, Sven and Lafarge, Florent and Alliez, Pierre},
  journal={ISPRS journal of photogrammetry and remote sensing},
  volume={90},
  pages={68--82},
  year={2014},
  publisher={Elsevier}
}

@inproceedings{mura2016piecewise,
  title={Piecewise-planar reconstruction of multi-room interiors with arbitrary wall arrangements},
  author={Mura, Claudio and Mattausch, Oliver and Pajarola, Renato},
  booktitle={Computer graphics forum},
  volume={35},
  number={7},
  pages={179--188},
  year={2016},
  organization={Wiley Online Library}
}

@article{verdie2015lod,
  title={LOD generation for urban scenes},
  author={Verdie, Yannick and Lafarge, Florent and Alliez, Pierre},
  journal={ACM Transactions on Graphics},
  volume={34},
  number={3},
  pages={30},
  year={2015},
  publisher={Association for Computing Machinery}
}

@article{chen2008architectural,
  title={Architectural modeling from sparsely scanned range data},
  author={Chen, Jie and Chen, Baoquan},
  journal={International Journal of Computer Vision},
  volume={78},
  pages={223--236},
  year={2008},
  publisher={Springer}
}

@inproceedings{schindler2011classification,
  title={Classification and reconstruction of surfaces from point clouds of man-made objects},
  author={Schindler, Falko and W{\"o}rstner, Wolfgang and Frahm, Jan-Michael},
  booktitle={2011 IEEE International Conference on Computer Vision Workshops (ICCV Workshops)},
  pages={257--263},
  year={2011},
  organization={IEEE}
}

@inproceedings{van2011shape,
  title={On the shape of a set of points and lines in the plane},
  author={Van Kreveld, Marc and Van Lankveld, Thijs and Veltkamp, Remco C},
  booktitle={Computer Graphics Forum},
  volume={30},
  number={5},
  pages={1553--1562},
  year={2011},
  organization={Wiley Online Library}
}

@article{pan2025building,
  title={Building LOD representation for 3D urban scenes},
  author={Pan, Shanshan and Zhang, Runze and Liu, Yilin and Gong, Minglun and Huang, Hui},
  journal={ISPRS Journal of Photogrammetry and Remote Sensing},
  volume={226},
  pages={16--32},
  year={2025},
  publisher={Elsevier}
}

@article{li2021feature,
  title={Feature-preserving 3D mesh simplification for urban buildings},
  author={Li, Minglei and Nan, Liangliang},
  journal={ISPRS Journal of Photogrammetry and Remote Sensing},
  volume={173},
  pages={135--150},
  year={2021},
  publisher={Elsevier}
}

@article{ochmann2016automatic,
  title={Automatic reconstruction of parametric building models from indoor point clouds},
  author={Ochmann, Sebastian and Vock, Richard and Wessel, Raoul and Klein, Reinhard},
  journal={Computers \& Graphics},
  volume={54},
  pages={94--103},
  year={2016},
  publisher={Elsevier}
}

@inproceedings{gao2022low,
  title={Low-poly mesh generation for building models},
  author={Gao, Xifeng and Wu, Kui and Pan, Zherong},
  booktitle={ACM SIGGRAPH 2022 Conference Proceedings},
  pages={1--9},
  year={2022}
}

@article{chen2023robust,
  title={Robust low-poly meshing for general 3d models},
  author={Chen, Zhen and Pan, Zherong and Wu, Kui and Vouga, Etienne and Gao, Xifeng},
  journal={ACM Transactions on Graphics (TOG)},
  volume={42},
  number={4},
  pages={1--20},
  year={2023},
  publisher={ACM New York, NY, USA}
}

@inproceedings{garland1997surface,
  title={Surface simplification using quadric error metrics},
  author={Garland, Michael and Heckbert, Paul S},
  booktitle={Proceedings of the 24th annual conference on Computer graphics and interactive techniques},
  pages={209--216},
  year={1997}
}

@incollection{cohen2004variational,
  title={Variational shape approximation},
  author={Cohen-Steiner, David and Alliez, Pierre and Desbrun, Mathieu},
  booktitle={ACM SIGGRAPH 2004 Papers},
  pages={905--914},
  year={2004}
}

@inproceedings{salinas2015structure,
  title={Structure-aware mesh decimation},
  author={Salinas, David and Lafarge, Florent and Alliez, Pierre},
  booktitle={Computer Graphics Forum},
  volume={34},
  number={6},
  pages={211--227},
  year={2015},
  organization={Wiley Online Library}
}

@article{rabbani2006segmentation,
  title={Segmentation of point clouds using smoothness constraint},
  author={Rabbani, Tahir and Van Den Heuvel, Frank and Vosselmann, George},
  journal={International archives of photogrammetry, remote sensing and spatial information sciences},
  volume={36},
  number={5},
  pages={248--253},
  year={2006},
  publisher={Citeseer}
}

@incollection{cgal:Shape-Detection,
  author = {Sven Oesau and Yannick Verdie and Cl{\'e}ment Jamin and Pierre Alliez and Florent Lafarge and Simon Giraudot and Thien Hoang and Dmitry Anisimov},
  title = {Shape Detection},
  publisher = {{CGAL Editorial Board}},
  edition = {{6.0.1}},
  booktitle = {{CGAL} User and Reference Manual},
  url = {https://doc.cgal.org/6.0.1/Manual/packages.html#PkgShapeDetection},
  year = 2024
}

@inproceedings{schnabel2007efficient,
  title={Efficient RANSAC for point-cloud shape detection},
  author={Schnabel, Ruwen and Wahl, Roland and Klein, Reinhard},
  booktitle={Computer graphics forum},
  volume={26},
  number={2},
  pages={214--226},
  year={2007},
  organization={Wiley Online Library}
}

@incollection{cgal:Alpha-Shapes,
  author = {Tran Kai Frank Da},
  title = {{2D} Alpha Shapes},
  publisher = {{CGAL Editorial Board}},
  edition = {{6.0.1}},
  booktitle = {{CGAL} User and Reference Manual},
  url = {https://doc.cgal.org/6.0.1/Manual/packages.html#PkgAlphaShapes2},
  year = 2024
}

@inproceedings{qi2017pointnet++,
  title={PointNet++: Deep Hierarchical Feature Learning on Point Sets in a Metric Space},
  author={Qi, Charles R and Su, Hao and Mo, Kaichun and Guibas, Leonidas J},
  booktitle={Advances in neural information processing systems},
  year={2017}
}

@article{gonzalez2010measurement,
  title={Measurement of areas on a sphere using Fibonacci and latitude--longitude lattices},
  author={Gonz{\'a}lez, {\'A}lvaro},
  journal={Mathematical geosciences},
  volume={42},
  pages={49--64},
  year={2010},
  publisher={Springer}
}

@incollection{cgal:atw-aabb-24b,
  author = {Pierre Alliez and St{\'e}phane Tayeb and Camille Wormser},
  title = {{2D} and {3D} Fast Intersection and Distance Computation},
  publisher = {{CGAL Editorial Board}},
  edition = {{6.0.1}},
  booktitle = {{CGAL} User and Reference Manual},
  url = {https://doc.cgal.org/6.0.1/Manual/packages.html#PkgAABBTree},
  year = 2024
}

@article{boykov2004experimental,
  title={An experimental comparison of min-cut/max-flow algorithms for energy minimization in vision},
  author={Boykov, Yuri and Kolmogorov, Vladimir},
  journal={IEEE transactions on pattern analysis and machine intelligence},
  volume={26},
  number={9},
  pages={1124--1137},
  year={2004},
  publisher={IEEE}
}

@book{cgal:eb-24b,
  title = {{CGAL} User and Reference Manual},
  author = {{The CGAL Project}},
  publisher = {{CGAL Editorial Board}},
  edition = {{6.0.1}},
  year = 2024,
  url = {https://doc.cgal.org/6.0.1/Manual/packages.html}
}

@inproceedings{willis2022joinable,
  title={Joinable: Learning bottom-up assembly of parametric cad joints},
  author={Willis, Karl DD and Jayaraman, Pradeep Kumar and Chu, Hang and Tian, Yunsheng and Li, Yifei and Grandi, Daniele and Sanghi, Aditya and Tran, Linh and Lambourne, Joseph G and Solar-Lezama, Armando and others},
  booktitle={Proceedings of the IEEE/CVF conference on computer vision and pattern recognition},
  pages={15849--15860},
  year={2022}
}

@inproceedings{yeshwanth2023scannet++,
  title={Scannet++: A high-fidelity dataset of 3d indoor scenes},
  author={Yeshwanth, Chandan and Liu, Yueh-Cheng and Nie{\ss}ner, Matthias and Dai, Angela},
  booktitle={Proceedings of the IEEE/CVF International Conference on Computer Vision},
  pages={12--22},
  year={2023}
}

@inproceedings{he2024windpoly,
  title={WindPoly: Polygonal Mesh Reconstruction via Winding Numbers},
  author={He, Xin and Lv, Chenlei and Huang, Pengdi and Huang, Hui},
  booktitle={European Conference on Computer Vision},
  pages={294--311},
  year={2024},
  organization={Springer}
}

@article{kazhdan2013screened,
  title={Screened poisson surface reconstruction},
  author={Kazhdan, Michael and Hoppe, Hugues},
  journal={ACM Transactions on Graphics (ToG)},
  volume={32},
  number={3},
  pages={1--13},
  year={2013},
  publisher={ACM New York, NY, USA}
}
}
\clearpage
\setcounter{page}{1}
\maketitlesupplementary

In this supplement, we present implementation details, analyze parameter sensitivity and ablation effects, demonstrate the method's generalizability, and discuss its limitations.
\section{Implementation Details}
\subsection{Pseudo-code of Planar Visibility Calculations}
\label{sec:ppvc}
This section mainly shows how to calculate the visibility ratio of planes. We model \( \alpha \)-shapes as uniformly luminous objects and the keypoints as the point lights.
We begin by introducing the notations used in this computation. Let:
\begin{itemize}[leftmargin=2em]
    \item $\mathcal{P} = (P_1, \dots, P_m)$ be the set of planar primitives;
    \item $\mathcal{A} = (A_1, \dots, A_m)$ be the set of \( \alpha \)-shape polygons;
    \item $\mathcal{K} = (K_1, \dots, K_m)$ be the set of keypoints of plane $P_i$, treated as point light sources;
    \item $\mathcal{D} = (\mathbf{d}_1, \dots, \mathbf{d}_{50})$ be a sequence of 50 unit direction vectors generated by Fibonacci sampling on the unit sphere, representing the directions of sampled light rays.
    \item $\mathcal{T}$ is a hierarchical collision detection structure built from $\mathcal{A}$, where each node is bounded by an Axis-Aligned Bounding Box (AABB).
    \item $\mathcal{N} = \{\mathbf{n}_1, \dots, \mathbf{n}_m\}$ be the set of unit surface normals, where each $\mathbf{n}_i$ is the outward-pointing normal to the supporting plane $P_i$. Further, for a highly visible plane, we expect the normal to point to the visible half of the facet — the side where light does not intersect the model. Since normals are ambiguous, the final direction is determined by votes from scattered rays. Specifically, for the highly visible plane primitive $P_i$, let \(\mathbf{n}_i\) denote the original normal direction. The number of votes supporting \(\mathbf{n}_i\) as the final direction is \(N_{ni}^{+}\), and those supporting \(\mathbf{-n}_i\) are \(N_{ni}^{-}\).

    \item $N_{h}$ represents the number of highly visible planes, which is 0 at the beginning. $N_{h}^{pre}$ represents $N_{h}$ corresponding to the last iteration.
    \item $\mathcal{V} = (v_1, \dots, v_m)$ be the output visibility ratio of planes.
    \item $\mathcal{Q} = (Q_1, \dots, Q_m)$ be the output original visible labels of each planes. If the visibility ratio $v_i$ is larger than 0.5, the plane $P_i$ is defined as original highly visible ($Q_i$ = 1), otherwise, barely visible ($Q_i$ = 0).
\end{itemize}

The calculation of a planar visibility ratio is an iterative optimization process. A ray is considered visible if it satisfies one of the following two conditions: i) the ray does not intersect the model $\mathcal{T}$; ii) if the plane where the ray first intersects the model $\mathcal{T}$ has been determined to be highly visible, and the ray lies on the visible side of this plane, then the ray is also considered visible. We iterate on the above process until the number of strong visible planes no longer increases.

\begin{algorithm}[H]
\caption{Plane visibility calculations}
\begin{algorithmic}[1]
\State \textbf{Require : }$\mathcal{A} \gets (A_1, \dots, A_m)$
\State \textbf{Require : }$\mathcal{K} \gets (K_1, \dots, K_m)$
\State \textbf{Require : }$\mathcal{D} \gets (\mathbf{d}_1, \dots, \mathbf{d}_{50})$
\State \textbf{Require : }$\mathcal{N} \gets \{\mathbf{n}_1, \dots, \mathbf{n}_m\}$
\State \textbf{Require : }$N_h \gets 0, N_{h}^{pre} \gets 0$
\State \textbf{Ensure : }$\mathcal{V} \gets (0, 0, \dots, 0)$
\State \textbf{Ensure : }$\mathcal{Q} \gets (0, 0, \dots, 0)$
\Statex
\Procedure{Plane visibility calculations}{}
    \State $\mathcal{T} \gets$ new \texttt{AABB\_Tree} from input geometry $\mathcal{A}$
    \Repeat
        \State $N_{h}^{pre} \gets N_h$
        \For{$P_i \in \mathcal{P}$ where $L_i = 0$}
            \State $C_i \gets 0,\quad C_n \gets 0$
            \For{$k_{ij} \in K_i$}
                \For{$d_t \in \mathcal{D}$}
                    \State Emit ray $r_t$ from $k_{ij}$ in direction $d_t$
                    \If{$r_t$ does not intersect in $\mathcal{T}$}
                        \State $C_n \gets C_n + 1$, \Call{Counts}{$\mathbf{n}_i, d_t$}
                    \Else
                        \State Let $p_k$ be the first intersection plane of $r_t$ with $\mathcal{T}$
                        \If{$Q_k = 1 \land \mathbf{n}_k \cdot d_t < 0$}
                            \State $C_n \gets C_n + 1$, \Call{Counts}{$\mathbf{n}_i, d_t$}
                        \Else
                            \State $C_i \gets C_i + 1$
                        \EndIf
                    \EndIf
                \EndFor                
            \EndFor
            \If{$C_i < C_n$}
                \State $Q_i \gets 1$, $N_h \gets N_h + 1$
                \If{$N_{ni}^{+} < N_{ni}^{-}$}
                    \State $\mathbf{n}_i \gets -\mathbf{n}_i$
                \EndIf
            \EndIf
            \State $v_i \gets C_i / (C_i + C_n)$
        \EndFor
    \Until{$N_{h}^{pre} = N_h$}
\EndProcedure
\Statex
\Procedure{Counts}{$\mathbf{n}_i, d_t$}
    \If{$\mathbf{n}_i \cdot d_t > 0$}
        \State $N_{ni}^{+} \gets N_{ni}^{+} + 1$
    \Else
        \State $N_{ni}^{-} \gets N_{ni}^{-} + 1$
    \EndIf
\EndProcedure
\end{algorithmic}
\end{algorithm}

\subsection{Pseudo-code of Missing Planes Recovery}
\label{sec:mpr}
The algorithm of this part is mainly divided into two steps. The first is the selection of singular line segments, and then is the planes fitting based on the singular line segments. Let:
\begin{itemize}[leftmargin=2em]
    \item $\mathcal{P} = (P_1, \dots, P_m)$ be the set of planar primitives;
    \item $\mathcal{L} = (L_1, \dots, L_m)$ be the set of intersection line sets, where each $L_i$ contains the intersection lines between $P_i$ and every other plane, i.e. $L_{ij} = P_i \cap P_j$ for $j \ne i$.
    \item $\mathcal{S} = (S_1, \dots, S_m)$ be the collection of boundary segment sets for each plane primitive, where:
        \begin{itemize}
        \item $S_i = (s_{i1}, \dots, s_{in_i})$ represents the boundary segments of plane primitive $P_i$ (with $n_i$ segments).
        \item $\mathbf{n}_{ij}$ denotes the normal vector of segment $s_{ij}$.
        \item $I_{ij}$ is the set of interior points of segment $s_{ij}$.
        \item $c_{ij}$ denotes the centroid of the line segment $s_{ij}$, which is calculated as the average position of all points contained in $I_{ij}$.
        \item $f_{ij}$ is the initial plane of segment $s_{ij}$, determined by point $c_{ij}$ and normal vector $\mathbf{n}_{ij}$.
        \item $N_{ij} = \{s_{pq}\}$ as the neighborhood set of boundary segment $s_{ij}$, where two segments are mutual neighbors if their corresponding plane primitives $P_i$ and $P_p$ are second-order adjacent (see Section 3.3).
        \item $d_{ij} = 0$ or $1$. $d_{ij} = 0$ indicates that the line segment $s_{ij}$ has not been visited yet, while $d_{ij} = 1$ means that the line segment has already been visited.
        \end{itemize}
    \item $r_d$: distance threshold, $r_a$: angle threshold
    \item $\mathcal{G}$ be the output set of singular segments (see Section 3.3).
    \item $\mathcal{N}$ be the output recovered missing planes.
\end{itemize}

\subsection{Pseudo-code of Hierarchical Space Partition}
\label{sec:hsp}
We design a hierarchical space partition strategy based on a kinetic data structure. Unlike the previously described scheme, we divide the planes into three different categories based on visibility (highly visible, barely visible, and invisible), corresponding to three levels. Planes at the current level will partition further based on the outcome only after planes at the immediately higher level have completed their spatial partition (i.e. hierarchical partition). At the same time, each plane is given a different growth speed according to its visibility. This hierarchical strategy produces lightweight and semantically meaningful 3D partitions, with substantially reduced algorithmic complexity compared to other methods. Let:
\begin{itemize}[leftmargin=2em]
    \item $\mathcal{C}_h = (C_{h1}, \dots, C_{hh})$ be the set of convex polygons of highly visible planes (planes on level 1). Let $V_1$ be the set of non-frozen vertices on level 1.
    \item $\mathcal{C}_b = (C_{b1}, \dots, C_{bb})$ be the set of convex polygons of barely visible planes (planes on level 2). Let $V_2$ be the set of non-frozen vertices on level 2.
    \item $\mathcal{C}_m = (C_{m1}, \dots, C_{mm})$ be the set of convex polygons of invisible planes (planes on level 3). Let $V_3$ be the set of non-frozen vertices on level 3.
    \item $s_i$ be the growth speed of vertex $v_i$, which is positively correlated with its planar visibility ratio (see Section 3.5).
    \item $\mathcal{M}$ be the output partition of polyhedra.
\end{itemize}

\begin{algorithm}[H]
\caption{Missing planes recovery}
\begin{algorithmic}[1]
\State \textbf{Require : } $\mathcal{P} \gets (P_1, \dots, P_m)$
\State \textbf{Require : } $\mathcal{L} \gets (L_1, \dots, L_m)$
\State \textbf{Require : } $\mathcal{S} \gets (S_1, \dots, S_m)$
\State \textbf{Require : } $r_d$: distance threshold, $r_a$: angle threshold
\State \textbf{Ensure : } $\mathcal{G} \gets \emptyset$
\State \textbf{Ensure : } $\mathcal{N} \gets \emptyset$
\Statex
\Procedure{Singular segments selection}{}
    \For{$P_i \in \mathcal{P}$}
        \For{$s_{ik} \in S_i$}
            \For{$L_{ij} \in L_i$}
                \If{\parbox[t]{\dimexpr\linewidth-4em}{%
                $r_d < \text{dist}(c_{ik}, L_{ij}) \lor r_a < \text{angle}(s_{ik}, L_{ij})$
                }}
                    \State Add $s_{ik}$ to $\mathcal{G}$
                \EndIf
            \EndFor
        \EndFor
    \EndFor  
\EndProcedure
\Statex

\Procedure{Plane fitting}{}
\State Sort the segments in $\mathcal{G}$ in descending order of length.
\State Initialize $d_j \gets 0$ for all $s_j \in \mathcal{G}$
    \For{$s_{ik} \in \mathcal{G}$ where $d_{ik} = 0$}
        \State Initialize empty queue $Q$
        \State $\text{count} \gets 1$
        \State $f_{fit} \gets f_{ik}$
        \State $Q.push(s_{ik})$
        \State $d_{ik} \gets 1$
        \While{$Q$ is not empty}
            \State $s_{current} \gets Q.front()$
            \State $Q.pop()$            
            \State $\text{neighbors} \gets N_{current}$
            \For{each $s_j \in \text{neighbors}$ where $d_j = 0$}
                \If{\parbox[t]{\dimexpr\linewidth-4em}{%
                                $\text{dist}(I_{j}, f_{fit}) < r_d \land \text{angle}(\mathbf{n}_{j}, f_{fit}) < r_a$
                                }}
                    \State $Q.push(s_j)$
                    \State $d_j \gets 1$
                    \State $\text{count} \gets \text{count} + 1$
                    \State $\text{update\_plane\_model}(f_{fit}, s_j)$ 
                \EndIf
            \EndFor            
        \EndWhile
        \If{$\text{count} > 1$}
            \State $\mathcal{N} \gets \mathcal{N} \cup \{f_{fit}\}$
        \EndIf
    \EndFor  
\EndProcedure
\end{algorithmic}
\end{algorithm}

\begin{algorithm}[H]
\caption{Hierarchical space partition}
\begin{algorithmic}[1]
\State \textbf{Require : }$\mathcal{C}_h \gets (C_{h1}, \dots, C_{hh})$
\State \textbf{Require : }$\mathcal{C}_b \gets (C_{b1}, \dots, C_{bb})$
\State \textbf{Require : }$\mathcal{C}_r \gets (C_{r1}, \dots, C_{rr})$
\State \textbf{Require: } For each vertex $v_i$ from all polygons, $s_i := \text{growth speed of } v_i$
\State \textbf{Ensure : }$\mathcal{M} \gets \emptyset$
\Statex
\Procedure{Hierarchical space partition}{}
    \State Compute the bounding box of all convex polygons $\{\mathcal{C}_h, \mathcal{C}_b, \mathcal{C}_m\}$
    \State Set the bounding box as $\text{level}\_0$ of the partition ($\text{space}\_0$)
    \State Generate the initial non-frozen vertices sets $\{\mathcal{V}_1, \mathcal{V}_2, \mathcal{V}_3\}$
    \For{$i \gets 1$ \textbf{to} $3$}
        \While{$\mathcal{V}_i$ is not empty}    
            \State Get the highest priority vertex $v_j$ from $\mathcal{V}_i$
            \State Grow $v_j$ with speed $s_j$ based on $\text{space}\_(i-1)$ 
            \State Determine the collision case (see Figure 6)
            \State Update $\mathcal{V}_i$ with sliding and/or frozen vertices
            \State Update the $\text{level}\_i$ of the partitions ($\text{space}\_i$)
        \EndWhile
    \EndFor  
    \State Assemble adjacent facets of $\text{level}\_3$ of the partitions ($\text{space}\_3$) into polyhedra $\mathcal{M}$
\EndProcedure
\end{algorithmic}
\end{algorithm}

\section{Parameter Sensitivity Analysis}
\label{sec:psa}

In the proposed multi-stage reconstruction pipeline, several key parameters are involved, including (i) the weighting coefficients for plane categorization, (ii) the distance and angular thresholds used for missing plane recovery, and (iii) the weights applied during surface extraction.
In this section, we analyze the sensitivity of these parameters and show that most of them are robust across diverse scenes, with only a few requiring scene-specific tuning.

\subsection{Weighting Parameter for Plane Categorization}
The weighting parameter $w$ balances the influence between a node’s own visibility evidence and the contextual information from its neighboring nodes during plane classification. Larger values of $w$ place greater emphasis on neighborhood consistency, while $w = 0$ degenerates the classification to a purely local decision based solely on the node’s visibility ratio. In contrast, overly large values (e.g., $w = 1$) may over-smooth the labeling by suppressing local evidence, which can negatively affect classification quality.

To evaluate the impact of $w$, we perform experiments on a validation set consisting of 10 indoor scenes selected from the ScanNet++V2 dataset. As ground-truth visibility labels (i.e., highly vs. barely visible planes) are not available, we adopt a semantic-based proxy derived from the dataset annotations. Specifically, planes belonging to structural elements (e.g., walls, floors, and ceilings) are treated as highly visible, while planes associated with non-structural objects (e.g., tables and chairs) are regarded as barely visible. This mapping serves as an empirical approximation rather than a strict correspondence; therefore, we evaluate semantic–visibility consistency instead of classification accuracy.

Let $\mathcal{P} = \{P_i\}_{i=1}^{N}$ denote the set of extracted planar regions. Each plane $P_i$ is associated with a predicted visibility label $V_i \in \{\text{highly visible}, \text{barely visible}\}$ and a semantic label $S_i$. We define two semantic category sets:
\begin{align} \mathcal{S}_{\text{struct}} &= \{\text{wall, floor, ceiling, \ldots}\}, \\ \mathcal{S}_{\text{non-struct}} &= \{\text{table, chair, \ldots}\}, \end{align}

A plane is considered consistent if its predicted visibility agrees with its semantic category. This is quantified by the indicator
\begin{equation} M_i = \mathbb{I}\!\left[ f(V_i) = g(S_i) \right], \end{equation}
where $f(\cdot)$ and $g(\cdot)$ map visibility and semantic labels to binary values, respectively. The overall consistency score is then computed as
\begin{equation}
C = \frac{1}{N} \sum_{i=1}^{N} M_i.
\end{equation}

This consistency metric captures the empirical alignment between visibility-based plane classification and widely observed semantic priors in indoor environments, and is used as an auxiliary measure to assess the stability of the proposed classification with respect to $w$.

\begin{figure}[htbp]
  \centering
  \includegraphics[width=0.99\linewidth]{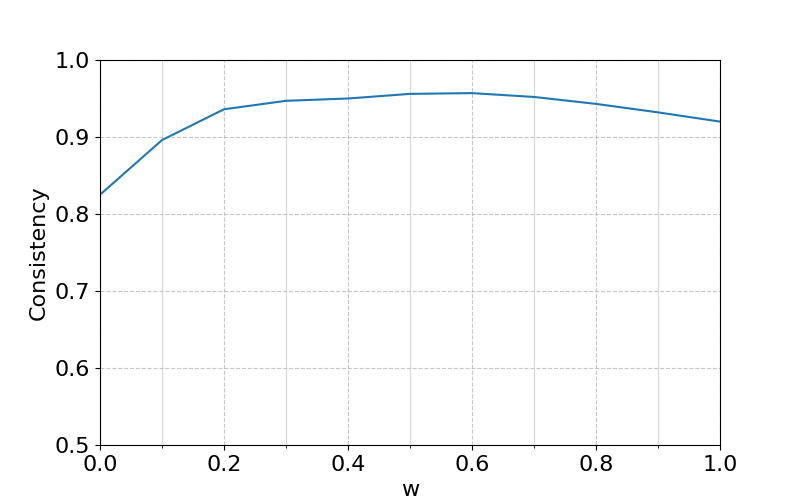}
  \caption{Impact of the parameter $w$. The consistency metric drops noticeably when $w$ approaches the extremes, while higher consistency is achieved when $w$ lies in the range of 0.4 to 0.7.}
  \label{fig:sp101}
\end{figure}

As shown in Figure~\ref{fig:sp101}, when $w$ varies from 0 to 1, the consistency score decreases at both extremes.
Visual inspection indicates that these regions correspond to noticeable classification errors.
In contrast, the consistency remains relatively high and stable when $w \in [0.4, 0.7]$.
Based on this observation, we set $w = 0.5$ in all experiments.

\subsection{Thresholds for Missing Plane Recovery}
The angle and distance thresholds in this module are used to extract singular segments, i.e., segments that cannot be sufficiently explained by existing planar structures, and to guide the fitting of corresponding missing planes. A larger threshold tends to classify more segments as being supported by existing planes, resulting in fewer singular segments and consequently fewer recovered missing planes. Conversely, an excessively small threshold may introduce a large number of false singular segments, leading to erroneous plane fitting and increased reconstruction error.

The angle threshold is determined following the strategy proposed in~\cite{han2021vectorized, fang2021structure}, while the distance threshold is empirically set in accordance with common practices in point-cloud-based plane fitting literature~\cite{ochmann2016automatic, xiang2024efficient}. Given that our test datasets exhibit relatively uniform point density and consistent structural characteristics, the selected thresholds demonstrate stable performance across different scenes.

To evaluate parameter sensitivity, we adopt a synthetic masking-and-recovery protocol on a dedicated validation set consisting of 50 models randomly selected from the ScanNet++ V2 dataset.
Due to the absence of ground-truth annotations for missing planes in real-world scans, we simulate partial observations by randomly removing a subset of existing planar structures along with their associated point sets.
These removed planes serve as the ground truth for recovery, denoted as $\mathcal{G} = \{G_j\}_{j=1}^{M}$.
Given the incomplete scans, our algorithm produces a set of recovered planes $\mathcal{R} = \{R_i\}_{i=1}^{K}$, which are then matched against $\mathcal{G}$ for quantitative evaluation.

To accommodate varying geometric complexity across different planes, we employ an adaptive matching criterion.
A recovered plane is regarded as a true positive if both its angular deviation and spatial distance to a ground-truth plane fall within twice the maximum fitting residual of the ground-truth plane’s supporting points.
A one-to-one matching constraint is enforced to prevent duplicate assignments.
Performance is evaluated using Precision ($\text{TP}/|\mathcal{R}|$), Recall ($\text{TP}/|\mathcal{G}|$), and the $F_1$-score:
\begin{equation}
    \text{F}_1 = \frac{2 \cdot \text{Precision} \cdot \text{Recall}}
    {\text{Precision} + \text{Recall}}.
\end{equation}

\begin{figure}[htbp]
  \centering
  \includegraphics[width=0.99\linewidth]{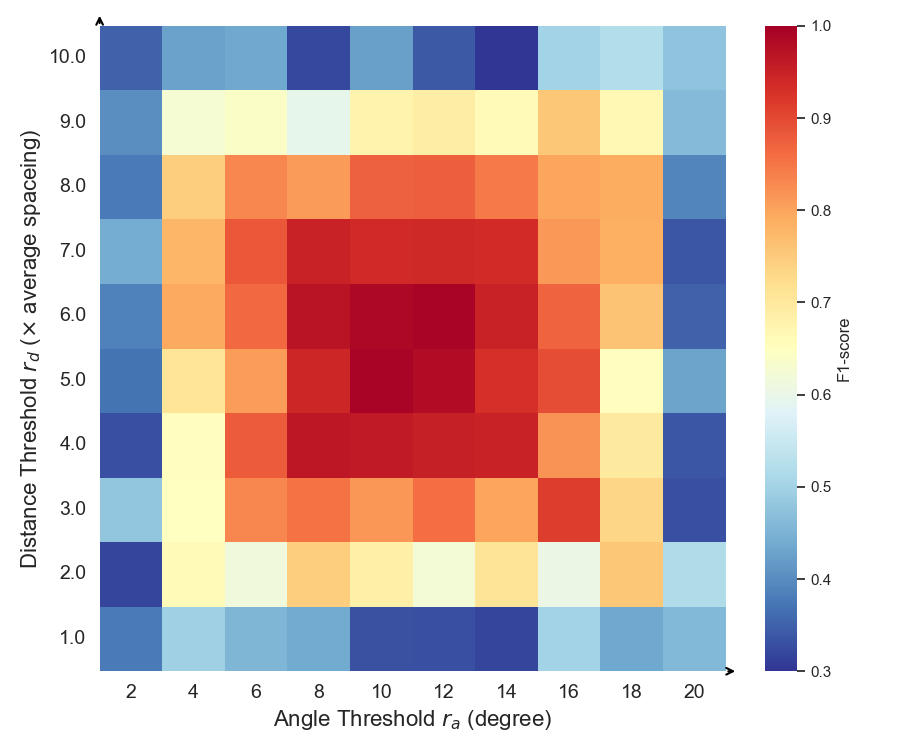}
  \caption{Impact of the angular threshold and the distance threshold.}
  \label{fig:sp102}
\end{figure}

As shown in Figure \ref{fig:sp102}, when the angular threshold is set too small, the $F_1$-score of plane recovery decreases significantly.
This is primarily because an excessive number of singular edges are extracted, leading to the reconstruction of many erroneous planes.
Conversely, overly large angular thresholds suppress the detection of genuine missing planes, resulting in reduced recall.
Empirically, the best performance is achieved when the angular threshold lies between 8° and 12°, and the distance threshold is set to approximately 4 to 6 times the average plane fitting residual, yielding consistently higher $F_1$-scores.

\subsection{Weighting Parameter for Surface Extraction}

\begin{figure}[htbp]
  \centering
  \includegraphics[width=0.99\linewidth]{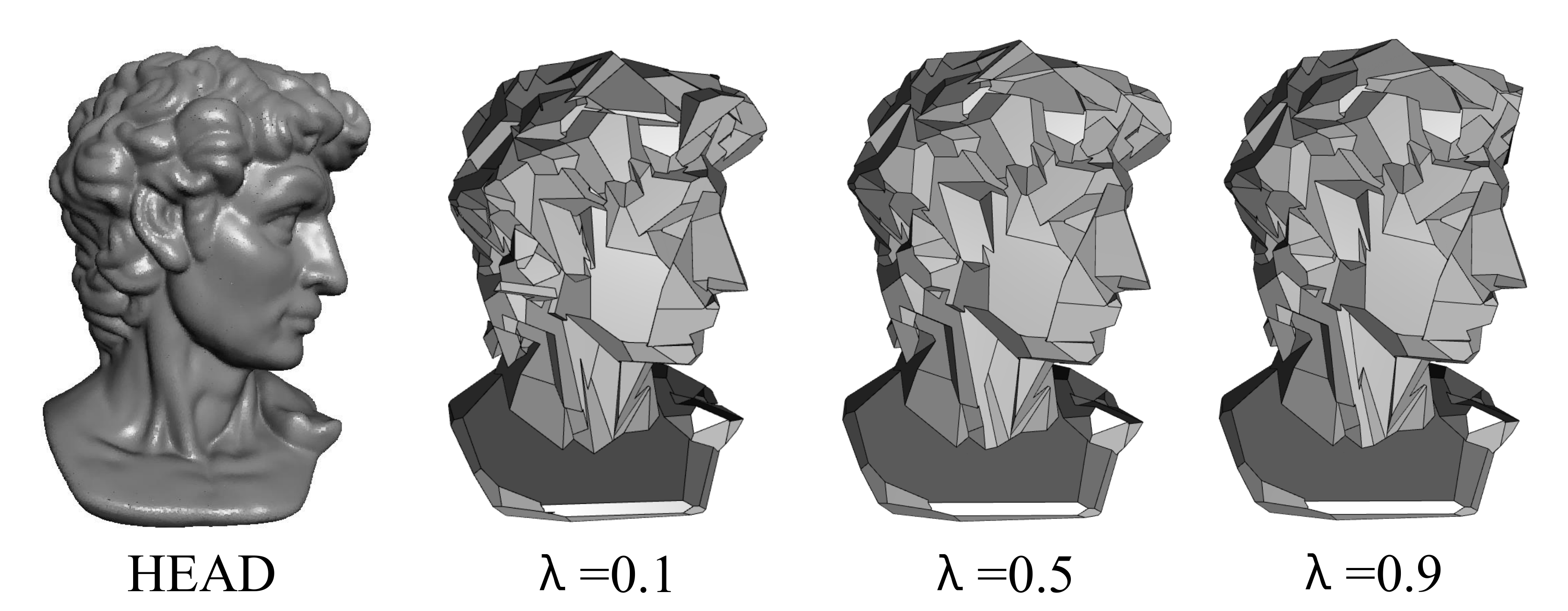}
  \caption{Impact of the parameter $\lambda$. When $\lambda$ is too small, more erroneous and redundant surfaces are preserved (e.g., $\lambda=0.1$). In contrast, too large values of $\lambda$ lead to missing surface patches (e.g., $\lambda=0.9$). Table~\ref{tab:lambda_for_se} presents more detailed values.}
  \label{fig:sp103}
\end{figure}

This parameter primarily governs the trade-off between model compactness and geometric fidelity. 
When assigned an excessively large value (e.g., greater than 0.9), the reconstruction tends to suffer from noticeable structural loss and increased geometric errors. 
Conversely, overly small values (e.g., below 0.2) lead to redundant fine-scale details and the emergence of zigzag artifacts on the surface.
As illustrated in Figure \ref{fig:sp103} and Table~\ref{tab:lambda_for_se}, selecting the parameter within the range of 0.4 to 0.7 achieves a favorable balance between surface simplicity and reconstruction accuracy.

\begin{table}[b]
\centering
\resizebox{0.5\textwidth}{!}{
\begin{tabular}{c ccccccccc}
\toprule
$\lambda$ & 0.1 & 0.2 & 0.3 & 0.4 & 0.5 & 0.6 & 0.7 & 0.8 & 0.9 \\
\midrule
\#f       &838  &821  &805  &792  &781  &773  &720  &695  & 631 \\
$e$       &0.412&0.396&0.382&0.371 &0.372 &0.375 &0.379 &0.381&0.567 \\
\bottomrule
\end{tabular}}
\caption{Effect of $\lambda$. \#f is the number of output facets, and $e$ is the mean Hausdorff error (MHE) from input points to the output model.}
\label{tab:lambda_for_se}
\end{table}

\section{Ablation study}
\label{sec:as}
To validate the effectiveness of the proposed Missing Planes Recovery and Hierarchical Partitions, we conduct ablation studies on the dataset \textit{Arch\text{-}100}, as reported in Table~\ref{tab:ablation_final}. As shown in the results, removing Missing Planes Recovery leads to a significant degradation in accuracy. This is primarily attributed to the fact that certain locally missing details cannot be accurately reconstructed.
On the other hand, the absence of Hierarchical Partitions results in increased complexity in spatial partitioning and reduced model conciseness.

\newcommand{\cmark}{\ding{51}}
\newcommand{\xmark}{\ding{55}}

\begin{table*}[!htbp]
\centering
\setlength{\tabcolsep}{6pt} 
\begin{tabular}{l c c c c c c c c c}
\toprule
\multirow{2}{*}{ID} & \multicolumn{2}{c}{\textit{Components}} & \multicolumn{2}{c}{\textit{Correctness}} & \multicolumn{3}{c}{\textit{Conciseness}} & \multicolumn{1}{c}{\textit{Efficiency}} \\
\cmidrule(lr){2-3} \cmidrule(lr){4-5} \cmidrule(lr){6-8} \cmidrule(lr){9-9}
 & MP-Recovery & H-Partitions & $\text{MHE} \downarrow$ & $\text{RMSE} \downarrow$ & $\text{P}^{\text{Avg.}} \downarrow$ & $\text{F}^{\text{Avg.}} \downarrow$ & $\text{RH}^{\text{Avg.}} \downarrow$ & T(s) $\downarrow$ \\
\midrule
1 & \xmark & \xmark & 1.955 & 0.041 & 279 & 385 & 0.0060 & \textbf{231} \\
2 & \cmark & \xmark & 1.073 & 0.031 & 329 & 523 & 0.0042 & 320 \\
3 & \xmark & \cmark & 1.930 & 0.039 & \textbf{251} & \textbf{359} & 0.0053 & 309 \\
4(ours) & \cmark & \cmark & \textbf{1.066} & \textbf{0.025} & 315 & 501 & \textbf{0.0036} & 327 \\
\bottomrule
\end{tabular}
\caption{Ablation study of MP-Recovery and H-Partitions on dataset \textit{Arch\text{-}100}. \textbf{MP-Recovery}: Missing Planes Recovery; \textbf{H-Partitions}: Hierarchical Partitions. Correctness is quantified using Mean Hausdorff Error (\textit{MHE}) and Root Mean Squared Error (\textit{RMSE}), reflecting how well the reconstructed polygonal planes align with the input points. Conciseness is evaluated based on the number of vertices ($\text{P}^{\text{Avg.}}$) and facets ($\text{F}^{\text{Avg.}}$) in the reconstructed mesh, as well as a composite metric ($\text{RH}^{\text{Avg.}}$) that combines the Hausdorff distance and a simplification ratio. Computational efficiency is assessed via runtime (\textit{T}). The best results are highlighted in \textbf{bold}.}
\label{tab:ablation_final}
\end{table*}

\begin{figure*}[b]
  \centering
  \includegraphics[width=0.99\linewidth]{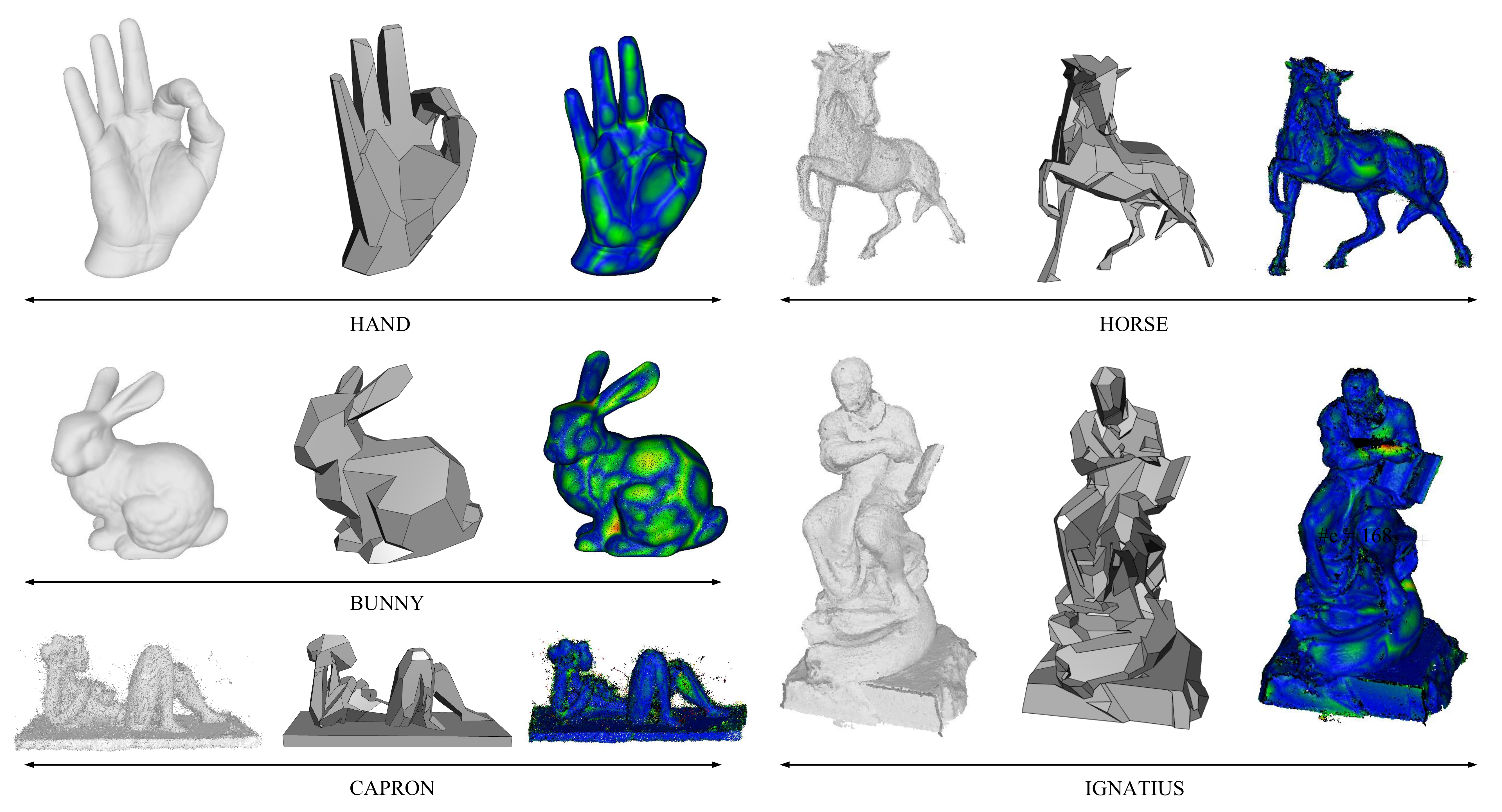}
  \caption{Reconstruction results on free-form objects. For each instance, we present the input point cloud with aligned normals (left), the generated watertight polygonal mesh (center), and a heatmap visualizing the RMSE (root-mean-square error) between the reconstructed model and the input point cloud (right).}
  \label{fig:sp105}
\end{figure*}

\section{Flexibility}
\label{sec:as}
To further demonstrate the flexibility and generalization capability of our method, we conduct additional experiments on a set of free-form objects. It is worth noting that the point clouds of these objects are obtained using different acquisition techniques. Specifically, models such as Horse, Ignatius, and Capron are reconstructed from point clouds generated by multi-view stereo (MVS), while Bunny and Hand are acquired using laser scanning.

\begin{table}[!htbp]
\centering
\resizebox{0.5\textwidth}{!}{
\begin{tabular}{llccccc}
\toprule
 &  & BUNNY & HAND & CAPRON & HORSE & IGNATIUS \\
\midrule
\multirow{3}{*}{ }
 & Type & Laser & Laser & MVS & MVS & MVS \\
 & \#i  & 146K  & 369K & 168K & 788K & 1.4M \\
 & \#s  & 101   & 75  & 152  & 275 & 298  \\
\midrule
\multirow{2}{*}{Ours}
 & $e$ & 0.463 & \textbf{0.411} & \textbf{0.259} & \textbf{0.237} & \textbf{0.201} \\
 & \#f   & 121  & 95           & \textbf{151}    & 365          & \textbf{430}  \\
\midrule
\multirow{2}{*}{KSR}
 & $e$   & \textbf{0.454} & 0.423        & 0.282 & 0.249        & 0.212 \\
 & \#f   & \textbf{111}  & \textbf{92}   & 156   & \textbf{351}  & 443  \\
\bottomrule
\end{tabular}}
\caption{Reconstruction results on free-form surface objects. \#i is the number of input points, \#s is the number of detected planes, \#f is the number of facets of the reconstruction model, and $e$ is the mean Hausdorff error (MHE) from input points to the output model. The best results are highlighted in \textbf{bold}.}
\label{tab:comparison_SP}
\end{table}

Despite the diversity in object shapes and data sources, our approach consistently produces accurate, high-quality reconstructions, yielding compact spatial partitions and concise mesh representations, as shown in Figure \ref{fig:sp105} and Table~\ref{tab:comparison_SP}. 
It should be noted, however, that for these free-form objects, the missing plane recovery module rarely detects valid missing planes in practice, as the underlying planar structure assumption is largely violated.

\section{Limitations}
\label{sec:li}
Because our missing plane fitting relies on the normal information of boundary segments, if the normal difference is too large, it cannot be recovered, for example, we cannot restore the missing top surface of a quadrangular frustum. Moreover, as the reconstruction framework is designed to produce watertight models, it is not suitable for completely open environments like a tennis court.

\end{document}